\documentclass[sn-mathphys]{sn-jnl}% Math and Physical Sciences Reference Style
\jyear{2021}%

\theoremstyle{thmstyleone}%
\theoremstyle{thmstyletwo}%

\theoremstyle{thmstylethree}%

\usepackage{subfigure}
\usepackage{graphics}
\usepackage{comment}

\begin{document}

\title{EfficientFace: An Efficient Deep Network with Feature Enhancement for Accurate Face Detection}


\author[1]{\fnm{Guangtao} \sur{Wang}}\email{202243023@njnu.edu.cn}

\author*[1]{\fnm{Jun} \sur{Li}}\email{lijuncst@njnu.edu.cn}

\author[2]{\fnm{Zhijian} \sur{Wu}}\email{52215903015@stu.ecnu.edu.cn}

\author[1]{\fnm{Jianhua} \sur{Xu}}\email{xujianhua@njnu.edu.cn}

\author[3]{\fnm{Jifeng} \sur{Shen}}\email{shenjifeng@ujs.edu.cn}

\author[4]{\fnm{Wankou} \sur{Yang}}\email{wkyang@seu.edu.cn}

\affil*[1]{\orgdiv{School of Computer and Electronic Information}, \orgname{Nanjing Normal University}, \orgaddress{\city{Nanjing}, \postcode{210023}, \country{China}}}

\affil[2]{\orgdiv{School of Data Science and Engineering}, \orgname{East China Normal University}, \orgaddress{\city{Shanghai}, \postcode{200062}, \country{China}}}

\affil[3]{\orgdiv{School of Electrical and Information Engineering}, \orgname{Jiangsu University}, \orgaddress{\city{Zhenjiang}, \postcode{212013}, \country{China}}}

\affil[4]{\orgdiv{School of Automation}, \orgname{Southeast University}, \orgaddress{\city{Nanjing}, \postcode{210096}, \country{China}}}


\abstract{In recent years, deep convolutional neural networks (CNN) have significantly advanced face detection. In particular, lightweight CNN-based architectures have achieved great success due to their low-complexity structure facilitating real-time detection tasks. However, current lightweight CNN-based face detectors trading accuracy for efficiency have inadequate capability in handling insufficient feature representation, faces with unbalanced aspect ratios and occlusion. Consequently, they exhibit deteriorated performance far lagging behind the deep heavy detectors. To achieve efficient face detection without sacrificing accuracy, we design an efficient deep face detector termed EfficientFace in this study, which contains three modules for feature enhancement. To begin with, we design a novel cross-scale feature fusion strategy to facilitate bottom-up information propagation, such that fusing low-level and high-level features is further strengthened. Besides, this is conducive to estimating the locations of faces and enhancing the descriptive power of face features. Secondly, we introduce a Receptive Field Enhancement module to consider faces with various aspect ratios. Thirdly, we add an Attention Mechanism module for improving the representational capability of occluded faces. We have evaluated EfficientFace on four public benchmarks and experimental results demonstrate the appealing performance of our method. In particular, our model respectively achieves 95.1\% (Easy), 94.0\% (Medium) and 90.1\% (Hard) on validation set of WIDER Face dataset, which is competitive with heavyweight models with only 1/15 computational costs of the state-of-the-art MogFace detector.}

\keywords{Face detection, feature enhancement, cross-scale feature fusion, Receptive Field Enhancement, Attention mechanism}



\maketitle

\section{Introduction}\label{sec1}

Face detection is one of the most fundamental tasks in computer vision. Since the pioneering work built on Haar features and Adaboost classifier\cite{r48}, significant progress has been made in face detection. In particular, deep convolutional neural network (CNN) has enormously advanced face detection and achieved unrivaled performance compared to conventional methods. In pursuit of high performance, a great majority of heavyweight face detectors such as MogFace\cite{r21}, AlnnoFace\cite{r32} and DSFD\cite{r14} have been proposed recently. Although they have achieved superior performance, these heavyweight models usually comprise complex structures with excessive number of parameters. For instance, the advanced DSFD detector is designed by building a deep network with 100M$+$ parameters costing 300G$+$ MACs. Therefore, both training and inference of the networks not only require high-performance platform but also cost expensive overhead, which is quite demanding in practical applications. For efficiency, massive efforts are devoted to designing lightweight face detectors, yielding a variety of lightweight models including EXTD\cite{r17}, YOLOv5n-0.5\cite{r18} and LFFD\cite{r41}. With the help of simple design, these lightweight models enjoy compact structure with much fewer parameters. However, with simplified and pruned network, the lightweight models sacrificing accuracy for efficiency reveal severely degraded performance compared to the heavyweight counterparts. We assume that current lightweight face detectors excessively pursue the lightweight design. Consequently, they fail to sufficiently capture the characteristics of the faces when handling insufficient feature representation, faces with unbalanced aspect ratios and severe occlusion. Thus, this makes difficult for the lightweight models to achieve satisfactory detection performance, and hinders their real-world applications. For example, lightweight YOLOv5n model  exhibits tremendous superiority in efficiency with 70$\times$ less parameters and 125$\times$ less MACs compared to DSFD, whereas the former is still far inferior to the latter with a significant performance drop of 10\%.

To achieve efficient detection without compromising accuracy, we propose a deep network termed EfficientFace in this study. Developed from EfficientNet\cite{r40}, our model includes three key modules for feature enhancement: Symmetrically Bi-directional Feature Pyramid Network (SBiFPN), Receptive Field Enhancement (RFE) and Attention module (AM). Firstly, in order to help high-level features acquire location information, we shorten the feature propagation pathway between the two adjacent feature layers and design a SBiFPN module for cross-scale feature fusion, such that the resulting feature maps encoding both high-level semantic information and low-level face location information can better capture faces with insufficient representation capability. Secondly, taking into account substantial amount of faces with unbalanced aspect ratios in real-world applications, we introduce the Receptive Field Enhancement module following SBiFPN into our framework, such that the variance in the ratios of human faces is considered and modeled. Finally, we employ attention module for detecting occluded faces. The attention module combines both spatial-aware and channel-aware attention mechanisms, and thus can better localize and detect face regions with improved representational ability. 

To summarize, our contributions in this study are threefold as follows:

\begin{itemize}
    \item We propose a new framework for efficient face detection termed EfficientFace with lower complexity and fewer parameters. Exhibiting superior performance to the lightweight models, EfficientFace achieves a competitive detection performance in comparison to some heavyweight face detectors.
    \item In EfficientFace, we incorporate three key modules for feature enhancement. To be specifically, we firstly design a Symmetrically Bi-directional Feature Pyramid network (SBiFPN) to facilitate the feature propagation from bottom layer to top layer, such that the resulting feature maps encode both rich semantic information and accurate face location information. Meanwhile, we introduce the Receptive Field Enhancement module to detect faces with unbalanced aspect ratios, while add an attention module for better characterizing occluded faces.
    \item Extensive experiments on four public face detection benchmarks suggest the promise of the proposed network with superior efficiency compared to the state-of-the-art heavyweight detectors.
\end{itemize}

The rest of this paper is structured as follows. After reviewing the related work in Section~\ref{sec2}, we introduce our proposed EfficientFace detector in Section~\ref{sec3}. Subsequently, extensive experiments are conducted in Section~\ref{sec4} and the paper is concluded in Section~\ref{sec5}.
\section{Related Work}\label{sec2}

\subsection{Deep Face Detector}\label{subsec21}

Benefiting from the success of deep models developed for general-purpose object detection\cite{r1,r2,r3,r4,r5}, significant progress has been made in face detection. In particular, a variety of heavyweight face detectors have been designed for achieving accurate face detection\cite{r24,r25,r26,r27,r28}. Based on the improved SSD\cite{r3} detector, a new face detector termed S$^{3}$FD is proposed by Zhang et al.\cite{r12}. It contains a novel anchor matching strategy which has become an important strategy commonly used in face detection research. Wang et al.\cite{r22} developed an anchor-level attention mechanism to deal with face occlusion. Meanwhile, SSH\cite{r19} removes the fully connected layer of the classification network and uses the feature pyramid instead of the image pyramid, reducing parameters and speeding up the operation. Then, Tang et al.\cite{r13} proposed a new context assisted single shot face detector using context information termed PyramidBox. Another advanced detector is DSFD\cite{r14} which includes three modules: Feature Enhancement Module (FEM), Progressive Anchor Loss function (PAI) and new Data Enhancement strategy. Believing that the balance of training samples is critical for accurate detection, Ming et al.\cite{r42} proposed a group sampling method to balance the number of samples in each group during the training process. Liu et al.\cite{r15} indicated that more than 80\% of correctly predicted bounding boxes are regressed from mismatched anchors (IoU between anchor and face is below a threshold). Therefore, they proposed HAMBox framework incorporating an online high-quality anchor mining strategy, which compensates the faces that do not match the anchor with high-quality anchors. In addition, Li et al.\cite{r16} introduced an Automatic and Scalable Face Detector termed ASFD that combines neural architecture search techniques with a new loss design. 

Although the above models achieve superior performance, they usually have complex structure and architecture with tremendous parameters. Therefore, they incur considerable resources and costs during both training and inference processes. In addition to the above deep heavy face detectors, designing lightweight models has emerged as a major line of research in face detection. Compared to the heavy models, the advantage of the lightweight counterparts manifests itself in the compressed structure with largely reduced parameters. Thus, they incur limited costs and facilitate practical deployment in real-world applications. One of the most representative lightweight models is YOLOv5Face\cite{r18}. It was developed to modify and optimize YOLOv5\cite{r55} for face detection by a series of mechanisms. For instance, YOLOv5Face adds additional branches for face keypoint detection, replaces the Focus layer with a stem block structure, and utilizes smaller convolution kernels in SPP. Another well-known lightweight architecture EXTD\cite{r17} is an iterative network sharing model for multi-stage face detection. It significantly reduces the number of parameters, while it achieves degraded detection performance. This also suggests the drawback of the current lightweight detectors, which indicates that they usually sacrificed detection performance in return for higher efficiency with compact structure and reduced parameters.

\subsection{Feature Pyramid Network}\label{subsec22}

In object detection, it is widely acknowledged that fusing multi-scale features is substantially beneficial for boosting detection performance. Lin et al.\cite{r33} first proposed feature pyramid network (FPN) for multi-scale feature fusion in object detection. Subsequently, PANet\cite{r34} improved FPN by adding a bottom-up network structure following the output feature layer of FPN. Zhao et al.\cite{r36} proposed M2det model in which MLFPN was designed to handle the problem that the feature map of FPN used for object detection contains single-layer information. Ghiasi et al.\cite{r35} proposed to adopt neural architecture search to design a new FPN named NAS-FPN. Although the searching process is costly and time-consuming, NAS-FPN shows excellent detection performance. Based on EfficientNet, Tan et al.\cite{r11} proposed an efficient detection framework termed EfficientDet in which a weighted bidirectional feature Pyramid Network (BiFPN) is developed to quickly fuse multi-scale features. Combining attention mechanism and FPN, Cao et al.\cite{r37} proposed an attention-guided Context Feature Pyramid Network (AC-FPN). Recently, Wang et al.\cite{r38} proposed AF-FPN structure by using Adaptive Feature Fusion and Receptive Field Module to enhance the expression of feature pyramid. Qiao et al.\cite{r39} proposed a novel feature pyramid structure called Recursive feature Pyramid (RFP) which achieves promising performance. Nowadays, designing effective FPN structure remains an open problem, since cross-scale feature fusion is considerably beneficial for a variety of vision tasks when there exist significant variances in object scale and resolution.

\section{EfficientFace Face Detector}\label{sec3}

 In this section, we will firstly introduce the framework of our proposed EfficientFace model. Afterwards, we will elaborate on primary modules within the architecture. More specifically, these modules include Symmetrically Bi-directional Feature Pyramid Network (SBiFPN) module in Section~\ref{subsec32}, Receptive Field Enhancement (RFE) module in Section~\ref{subsec33}, and Attention Mechanism module(AM) in Section~\ref{subsec34}. In addition, the loss function is introduced in Section~\ref{subsec35}.

\subsection{The Network Architecture}\label{subsec31}

\begin{figure}
\centering
\includegraphics[width=100mm]{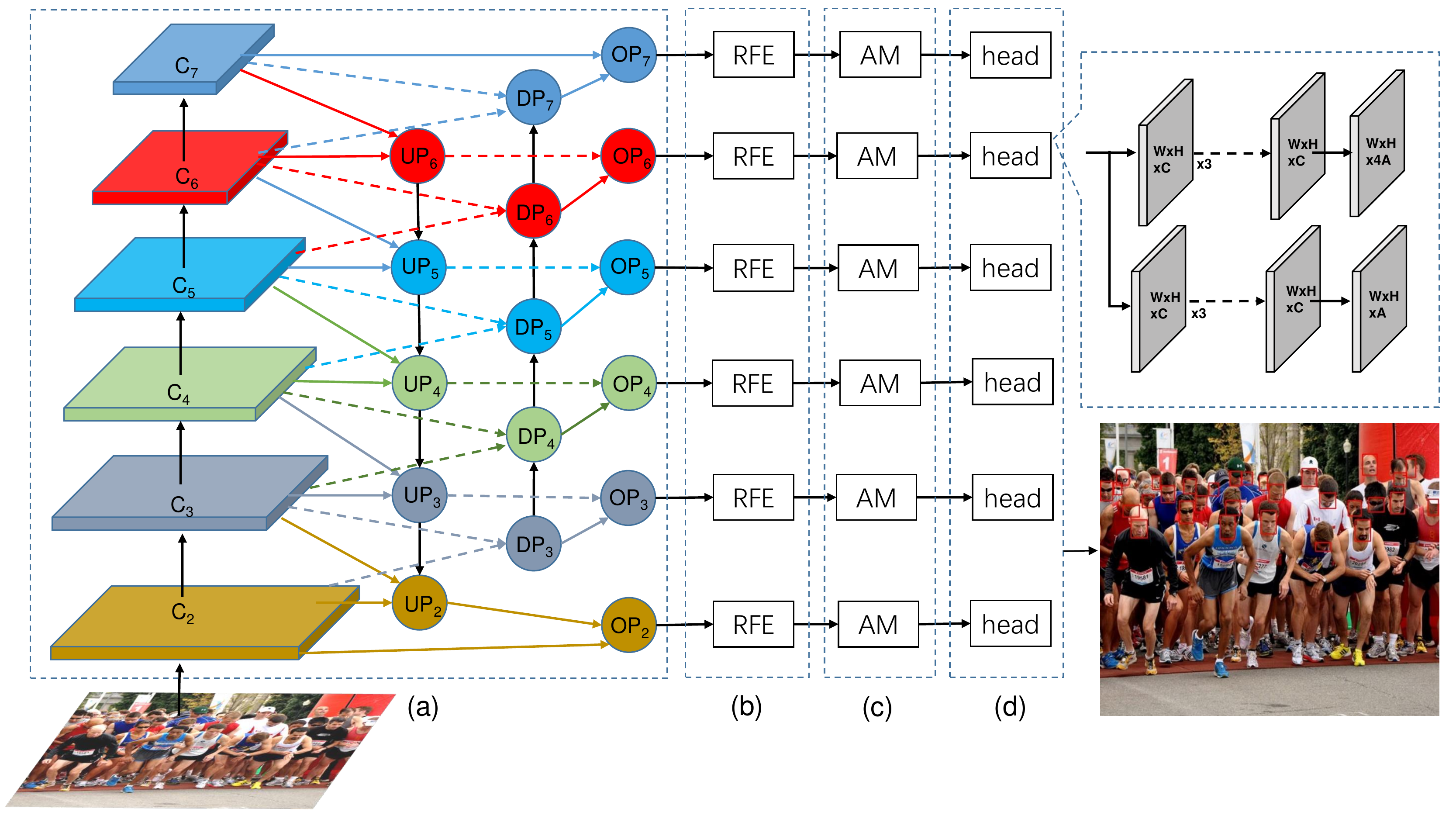}
\caption{The complete architecture of the proposed EfficientFace includes: the feature extraction network comprising backbone and SBiFPN (a), RFE module (b), AM module (c), and detection head with face classification and location network (d).}
\label{fig1}
\end{figure}

The network architecture of the proposed EfficientFace is shown in Figure~\ref{fig1}. With EfficientNet-B5 used as our backbone in EfficientFace, $C_{2}$, $C_{3}$, $C_{4}$ and $C_{5}$ are the feature maps extracted from the backbone, while $C_{6}$ and $C_{7}$ are obtained by downsampling $C_{5}$ and $C_{6}$ respectively. The downsampling factor is set to 2 in our network. The pathway from $UP_{6}$ to $UP_{2}$ denotes the top-down feature propagation pathway of FPN, and the counterpart from $DP_{3}$ to $DP_{7}$ is the added bottom-up propagation pathway. The two pathways function in parallel and constitute our SBiFPN. The pyramid structure is followed by the Receptive Field Enhancement module and the Attention Mechanism module.

\subsection{SBiFPN}\label{subsec32}

\begin{figure}
\centering
\includegraphics[width=100mm]{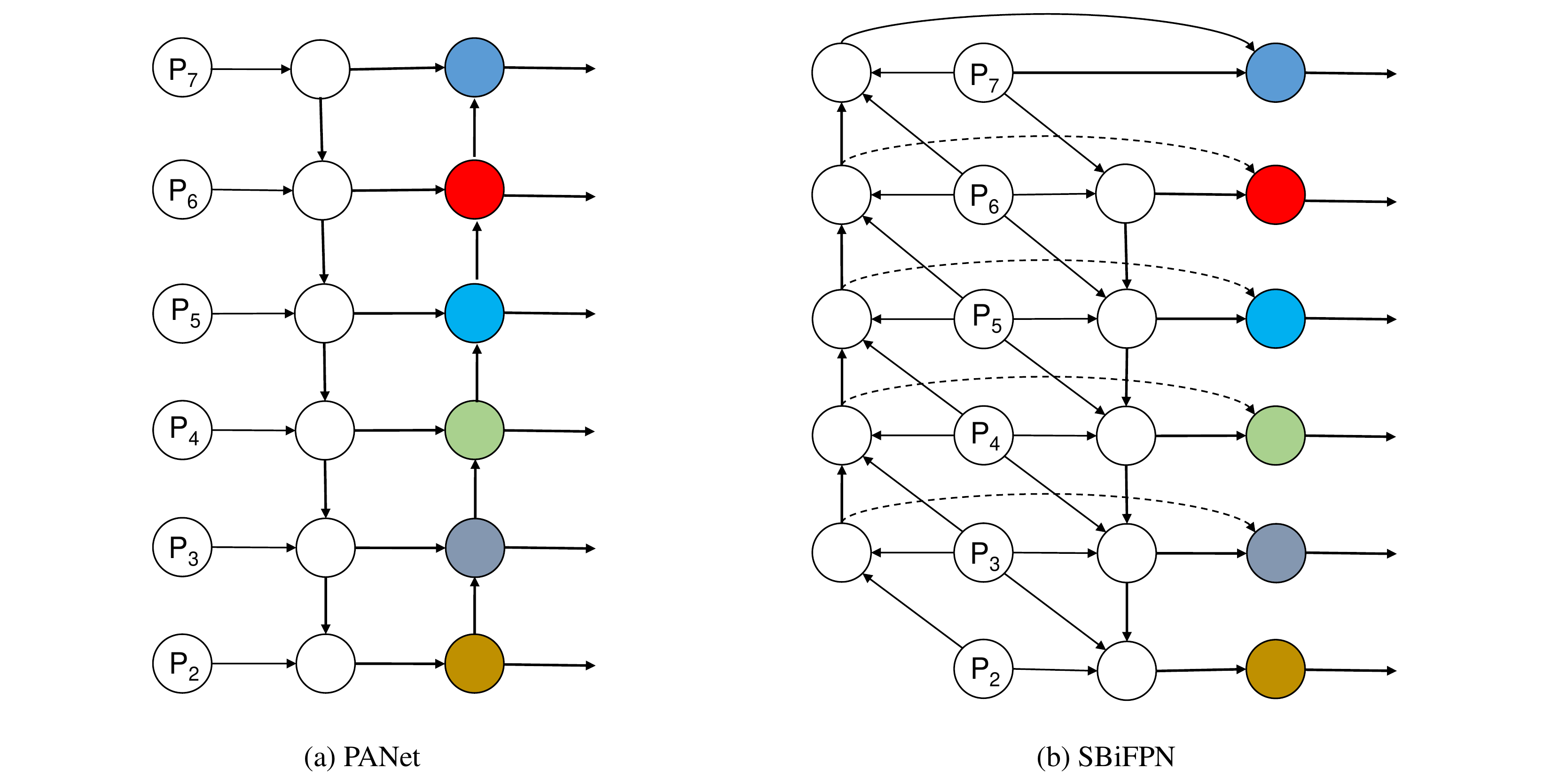}
\caption{Comparison of two feature networks. (a) In PANet, a bottom-up propagation pathway is added following FPN. (b) SBiFPN contains a parallel bottom-up pathway which is essentially symmetrical with FPN w.r.t. the backbone. The feature maps derived from FPN and the bottom-up pathway are combined for cross-scale feature fusion.}
\label{fig2}
\end{figure}

In the traditional FPN architectures, the feature propagation from low-level to top-level features passes through dozens of convolutional layers, making that top-level features fail to encode accurate face location due to the long-distance pathway. In order to mitigate this problem, we shorten the feature propagation distance between the two adjacent feature layers by designing a Symmetrically Bi-directional Feature Pyramid termed SBiFPN for cross-scale feature fusion. As shown in Fig.~\ref{fig2}, we impose a downsampling operation on each feature map $C_{i}$ and $DP_{i}$ at $i_{th}$ level respectively for scaling the feature map to half of the original. Then, feature map $C_{i+1}$ at next level is added to the two feature maps downsampled from $C_{i}$ and $DP_{i}$. The fused feature maps pass through a 3$\times$3 convolution kernel to generate feature maps $DP_{i+1}$. Finally, we fuse the generated feature maps $UP_{i+1}$ and $DP_{i+1}$, and obtain the resulting feature map $OP_{i+1}$ through a 3$\times$3 convolution layer. Since the bi-directional feature propagations are performed in parallel and they are essentially symmetrical with respect to backbone, it shortens the information flow path between low-level and high-level features. We assume the resulting feature maps simultaneously encode high-level semantics and low-level location information, and further enhance the fusion of multi-level cross-scale features.

Analogous to SBiFPN, both PANet and BiFPN perform bi-directional cross-scale feature fusion in feature pyramid network. In contrast to PANet and BiFPN where additional bottom-up pathway follows FPN, however, the bottom-up pathway functions in parallel with top-down counterpart in FPN within our network. Thus, the bi-directional propagation starts from the backbone simultaneously, implying that feature reuse is involved in our SBiFPN. As indicated in \cite{r59}, feature reuse is substantially beneficial for the success of convolutional neural networks. Comparison of our SBiFPN structure and classic PANet is illustrated in Fig.~\ref{fig2}.

Unlike the classic BiFPN structure, notably, we do not repeatedly iterate SBiFPN which performs only once to further make our EfficientFace network compact. Regarding the cross-scale fusion strategy, we follow BiFPN to perform fast weighted fusion method to aggregate multi-scale features \cite{r11}. Mathematically, the above SBiFPN fusion method can be expressed as the following three processes:

\begin{equation}
P_{7} = C_{7},P_{6} = C_{6},P_{5} = F(C_{5}),......,P_{2} = F(C_{2})
\end{equation}

(1) Top-down feature fusion process is formulated as follows:

\begin{equation}
UP_{i}= \left \{
\begin{array}{ll}%ll按顺序是公式左对齐和条件左对齐
    F(\alpha _{1} * P_{i} + \alpha _{2} * U(P_{i+1}) ), & i=6\\
    F(\alpha_{1} * P_{i} + \alpha_{2} * U(P_{i+1}) + \alpha_{3} * U(UP_{i+1}) ).         & i\in \left [2,5  \right ] 
\end{array}
\right.
\end{equation}

(2) Bottom-up feature fusion process is expressed as follows:

\begin{equation}
DP_{i}= \left \{
\begin{array}{ll}%ll按顺序是公式左对齐和条件左对齐
    F(\beta_{1} * P_{i} + \beta_{2} * P_{i-1} ), & i=3\\
    F(\beta_{1} * P_{i} + \beta_{2} * D(P_{i-1}) + \beta_{3} * D(DP_{i-1}) ).
    & i\in \left [4,7  \right ] 
\end{array}
\right.
\end{equation}

(3) With the above two processes completed, final feature fusion strategy is formulated as follows to generate the fused results:

\begin{equation}
OP_{i}= \left \{
\begin{array}{ll}%ll按顺序是公式左对齐和条件左对齐
    F(\gamma_{1} * P_{i} + \gamma_{2} * DP_{i}), 
    & i \in \left \{ 2,7 \right \} \\
    F(\gamma_{1} * UP_{i} + \gamma_{2} * DP_{i}).
    & i\in \left [3,6  \right ] 
\end{array}
\right.
\end{equation}

where $F(\cdot)$ denotes the convolution operation, $U(\cdot)$ represents the upsampling operation while $D(\cdot)$ means the down-sampling operation. In addition, $ \alpha, \beta, \gamma $ are the weights of the three groups of features fusion.

\subsection{Receptive Field Enhancement}\label{subsec33}

\begin{figure}
\centering
\includegraphics[width=110mm]{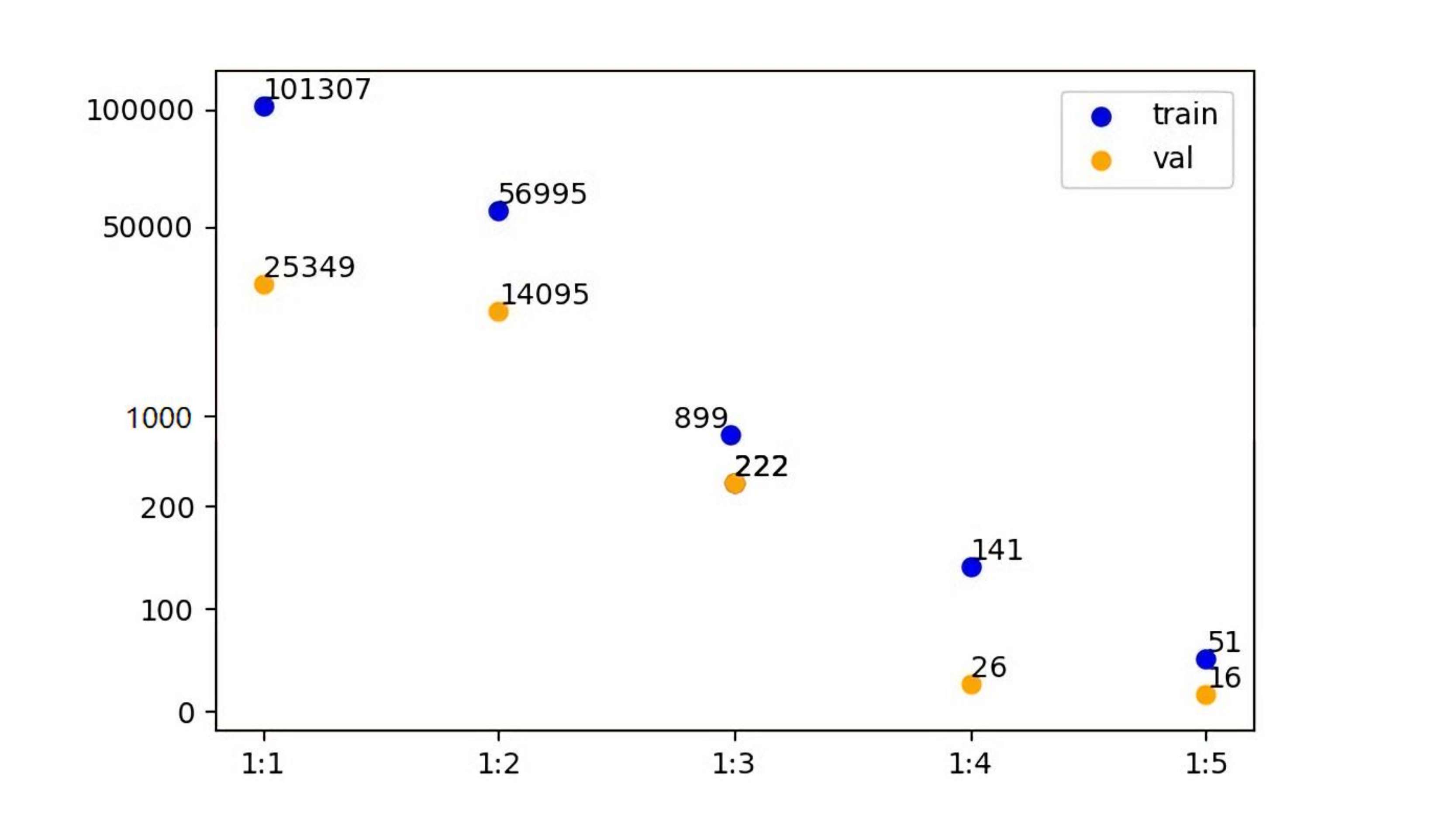}
\caption{A statistics of faces with varying aspect ratios in training and validation set on WIDER Face dataset. The abscissa represents the aspect ratio of face, and the ordinate represents the number of face.}
\label{fig3}
\end{figure}

It is well known that faces usually have unbalanced aspect ratios in images captured in real-world scenarios. For instance, as shown in Fig.~\ref{fig3}, a comprehensive statistics of faces with different aspect ratios on WIDER Face dataset\cite{r23}  suggests the aspect ratio of most faces is close to 1:1, while there still exist a large number of faces which are approximately 1:3 or 3:1. In some cases, the faces are even severely distorted with even 5:1 or 1:5 proportions. In order to alleviate this problem, we introduce a Receptive Field Enhancement (RFE) module\cite{r20} following SBiFPN.

\begin{figure}
\centering
\includegraphics[width=110mm]{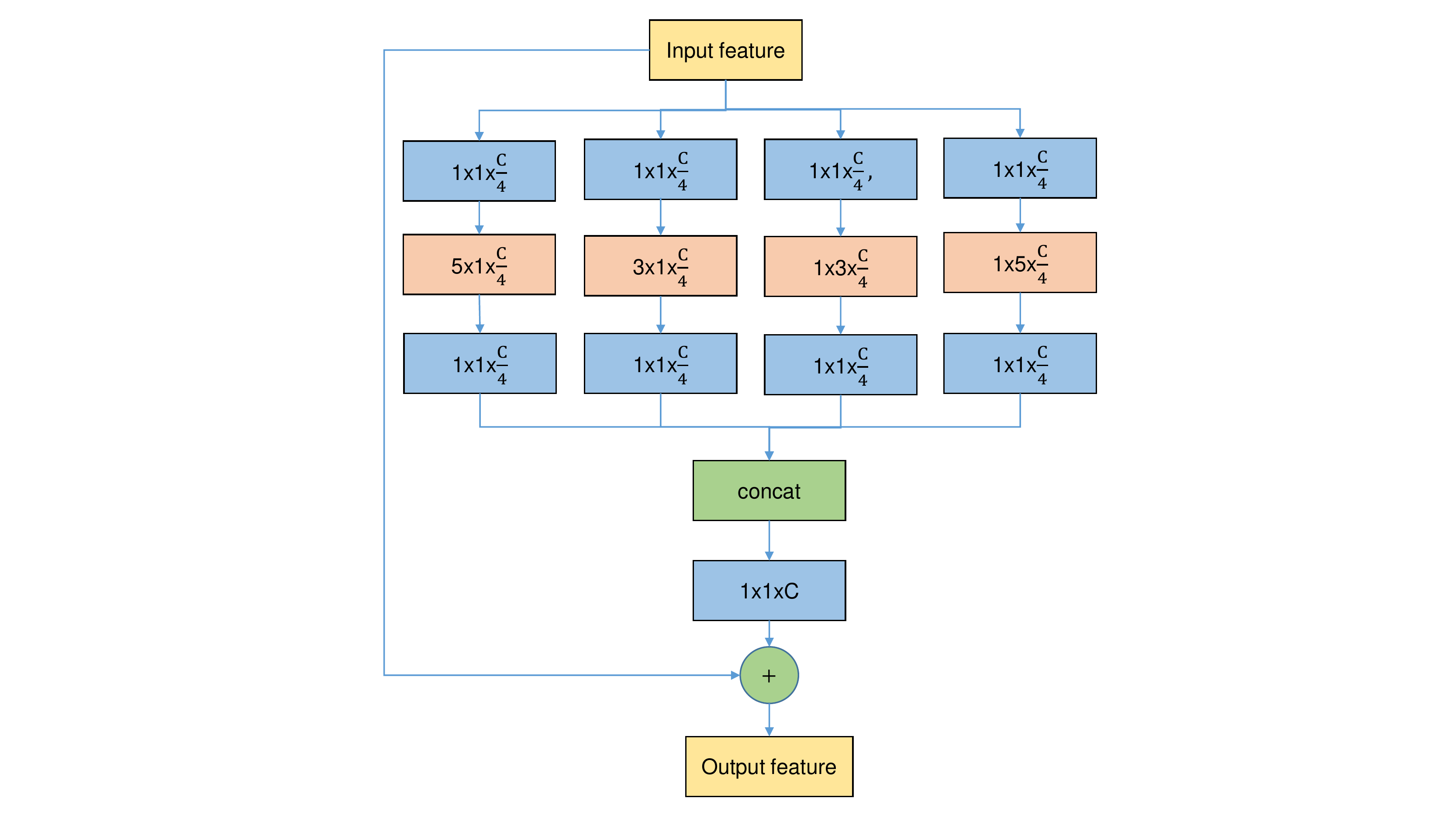}
\caption{Structure of the introduced RFE module.}
\label{fig4}
\end{figure}

The specific structure of RFE is illustrated in Fig.~\ref{fig4}. The input feature map is processed by four 1$\times$1 convolutional layers simultaneously for dimension reduction, and then they respectively pass through the following 1$\times$5, 1$\times$3, 3$\times$1 and 5$\times$1 convolutional layers. Finally, the resulting feature maps of each branch passing through the 1$\times$1 convolutional layer are concatenated and added to the input feature maps. The final output of module will have various receptive fields and can well handle the problem when there exist tremendous variances in the aspect ratio of faces.

\subsection{Attention Mechanism}\label{subsec34}

In face detection, occluded face makes only partial regions observed and lead to biased features, which prevents face detectors from achieving accurate detection. In order to mitigate this problem, we add an attention module\cite{r60} after the Receptive Field Enhancement module, such that occluded faces can be detected by identifying and enhancing the critical regions in the image.

\begin{figure}
\centering
\includegraphics[width=110mm]{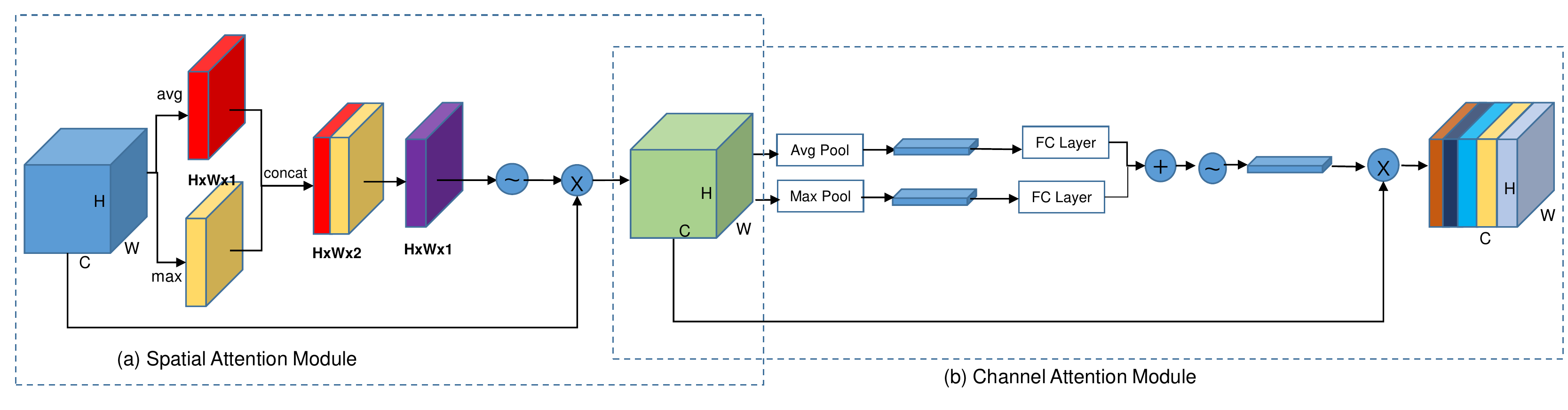}
\caption{Structure of Attention module in our EfficientFace. It consists of two consecutive components, namely spatial attention module and channel attention module.}
\label{fig5}
\end{figure}

In our EfficientFace, as shown in Fig.~\ref{fig5}, the attention module is divided into two consecutive components of spatial attention and channel attention. The spatial attention module allows our network to focus on task-related area and the channel attention module can discover the channels with important significance. Both of them are helpful for our network to accurately localize and classify occluded face. In addition, we also explore the depth of attention module in our experiments, and reveal that our model achieves the best detection performance when it is set to 2.

\subsection{Loss function}\label{subsec35}
The loss function of our EfficientFace model consists of two parts, one is used for computing classification accuracy while the other for estimating regression error of face localization. Taking into account the problem of sample imbalance, we leverage focal loss\cite{r43} for the classification loss function in Eq.~\ref{eq6}. Meanwhile, Smooth $\l_1$ loss is used for regression loss function to localize faces as shown in Eq.~\ref{eq8}. Mathematically, it is formulated as Eq.~\ref{eq5}:

\begin{equation}
\label{eq5}
L_{ef}=L_{focal}+\lambda * L_{smooth}
\end{equation}

where

\begin{equation}
\label{eq6}
L_{focal}=-\alpha_{t}(1-p_{t})^{\gamma}log(p_{t})
\end{equation}

and

\begin{equation}
p_{t}= \left \{
\begin{array}{ll}%ll按顺序是公式左对齐和条件左对齐
    p, & y=1 \\
    1-p. & otherwise 
\end{array}
\right.
\end{equation}

\begin{equation}
\label{eq8}
smooth_{L_{1}}(x)= \left \{
\begin{array}{ll}%ll按顺序是公式左对齐和条件左对齐
    0.5*x^{2}, & \lvert y \rvert <1 \\
    \lvert y \rvert -0.5. & otherwise
\end{array}
\right.
\end{equation}

In Eq.~\ref{eq5}, $\lambda$ is a parameter balancing the classification and regression loss. $p\in \left [0,1  \right ]$ is the probability estimated for the class with label 1, and $\alpha_{t}$ is the balancing factor. Besides, $\gamma$ is the focusing parameter that adjusts the rate at which simple samples are down-weighted. In implementation, we respectively set the values of $\lambda$, $\alpha_{t}$ and $\gamma$ to 1, 0.25 and 2.0\cite{r43}. 

\section{Experiments}\label{sec4}
In this section, we firstly introduce four public datasets where EfficientFace is evaluated in Section~\ref{subsec41}. Then, we discuss implementation details of our model in Section~\ref{subsec42}. Finally, experimental results and model analyses are presented in Section~\ref{subsec43}.

\subsection{Dataset}\label{subsec41}

In order to verify the efficacy of the proposed model, we have evaluated our EfficientFace network on four public benchmarking datasets for face detection. The datasets involved in our experiments are summarized as follows:
\begin{itemize}
    \item \textbf{AFW dataset}\cite{r44}. Released as an early face detection dataset, it has a total of 205 images and 473 labeled faces, demonstrating complex background and significant variances in faces.
    \item \textbf{Pascal Face dataset}\cite{r45}. It is a subset of the Pascal VOC dataset which is usually used for general-purpose object classification. The dataset consists of 851 images with 1,335 labeled faces.
    \item \textbf{FDDB dataset}\cite{r46}. It contains 2845 images with 5171 faces which are annotated with ellipses and rectangles. All the images are divided into grayscale and color images, while the dataset demonstrates a variety of challenges including difficult poses, low resolution and out-of-focus faces.
    \item \textbf{WIDER Face dataset}\cite{r23}. Known as the most challenging large-scale face detection dataset thus far, it is made up of 32,203 images with 393,703 annotated faces, which exhibit dramatic variances in face scales, occlusion, and poses. The dataset is split into training (40\%), validation (10\%) and testing sets (50\%). According to the detection rate of the EdgeBox\cite{r47}, the WIDER Face dataset is divided into three subsets depending on different difficulty levels of face detection, namely Easy, Medium and Hard subsets.
\end{itemize}

\subsection{Implementation Details}\label{subsec42}
In this section, we will discuss implementation details of the proposed EfficientFace model. We leverage EfficientDet for our baseline which is  pre-trained on COCO dataset. In particular, a $C_{2}$ layer is added to the EfficientDet network to detect small-size faces, whilst the anchor sizes used in the model are empirically set as \{16, 32, 64, 128, 256, 512\}. In addition, we use AdamW algorithm for network optimization and ReduceLROnPlateau attenuation strategy to adjust the learning rate which is initially set to $10^{-4}$. If the loss function stops descending within three epochs, the learning rate will be decreased by 10 times and eventually decay to $10^{-8}$. In SBiFPN module, the depth is set as 1 to avoid an excessive number of parameters resulting from iterative feature fusion network and reduce hardware configuration requirements. In addition, the maximum number of channels and the batch size of the EfficientFace network are empirically set to 288 and 4. The training and inference process are completed on a server equipped with a NVIDIA GTX3090 and PyTorch framework.

\subsection{Model Analysis}\label{subsec43}
In this section, we will carry out ablation studies in our network to explore the effect of individual modules on the performance of our model. Besides, a comprehensive comparative study will be conducted to compare our method with current state-of-the-art face detectors. All these experiments are conducted on WIDER Face dataset.

\subsubsection{The effects of multi-scale Feature Fusion}\label{subsec431}
Table~\ref{tab1} presents the performance of different multi-scale fusion networks with EfficientNet-B4 used as backbone. For fairness, the number of iterations of the fusion network is unanimously set to one, while the same weighted feature fusion strategy is utilized for different networks in our experiments. Since BiFPN can be treated as a simplified version of FPN$+$PANet, it reports slightly inferior performance compared to FPN$+$PANet as shown in Table~\ref{tab1}. Meanwhile, our proposed SBiFPN consistently beats BiFPN and FPN$+$PANet, achieving the highest AP scores at 94.4\%, 93.4\% and 89.1\% respectively on Easy, Medium and Hard subsets. In particular, SBiFPN outperforms the other two competitors by roughly 2\% on the Hard subset, demonstrating significant performance advantage. This suggests that the beneficial effect of facilitating the feature propagation between low-level and high-level features by shortening the bi-directional pathway with a symmetrical and parallel structure in SBiFPN. Consequently, the features generated from our SBiFPN enjoy superior representation capability.

\begin{table}[h] \addtolength{\textwidth}{0.0cm} \renewcommand{\arraystretch}{1.1}
\begin{center}
\caption{Comparison of different multi-scale feature fusion networks on WIDER Face dataset.}
\label{tab1}
\setlength{\tabcolsep}{8mm}{
\begin{tabular}{|c|c|c|c|}
\hline
Feature Networks & Easy & Medium & Hard\\
\hline
FPN$+$PANet & 93.0\%   & 91.7\%  & 87.2\%  \\
BiFPN     & 92.9\%   & 91.5\%  & 86.9\%  \\
SBiFPN    & \textbf{94.4\%}  & \textbf{93.4\%}  & \textbf{89.1\%}  \\
\hline
\end{tabular}}
\end{center}
\end{table}

\subsubsection{Influence of attention depth}\label{subsec432}
In this section, we append the Attention module to baseline detector which also adopts EfficientNet-B4 as backbone and test on single scale to explore the effect of depth of the attention module (number of attention modules conducted). As shown in Table~\ref{tab2}, inferior performance is observed compared to baseline when the attention module is performed only once. Conversely, when the attention module is conducted repeatedly, overall improved performance can be observed. In particular, highest detection accuracies are reported on all the three subsets when the attention modules are implemented twice. Excessively conducting the attention module repeatedly leads to a slight decline in the detection performance, incurring higher model complexity and additional parameters. Thus, we set the depth of the attention module as two in our experiments.

\begin{table}[h] \addtolength{\textwidth}{0.0cm} \renewcommand{\arraystretch}{1.1}
\begin{center}
\caption{Performance of the attention module with varying depth ($d$) on WIDER Face dataset.}
\label{tab2}
\setlength{\tabcolsep}{8mm}{
\begin{tabular}{|c|c|c|c|}
\hline
          & Easy & Medium & Hard\\
\hline
Baseline  & 93.8\%   & 92.5\%  & 87.2\%  \\
$d=1$         & 93.5\%   & 92.1\%  & 86.4\%  \\
$d=2$        & 93.9\%   & 92.8\%  & \textbf{88.0\%}  \\
$d=3$        & \textbf{94.3\%}   & \textbf{93.0\%}  & 87.6\%  \\
$d=4$         & 94.2\%   & 93.0\%  & 87.2\%  \\
$d=5$         & 93.4\%   & 92.4\%  & 87.4\%  \\
\hline
\end{tabular}}
\end{center}
\end{table}

\subsubsection{Effects of different modules}\label{subsec433}
In order to better study the influence of each module in our model, we further analyze it through ablation experiments. With EfficientNet-B5 used as backbone and SBiFPN incorporated, the detector reports respective 94.2\%, 93.6\%, and 89.7\% AP scores on Easy, Medium and Hard subsets as shown in Table~\ref{tab3}. When the RFE module is embedded following SBiFPN, the AP scores of our model are improved to 95.1\%, 93.9\% and 89.8\% respectively. Our complete model provides further performance improvement and reports the highest accuracy scores when all the three modules are incorporated into the network, achieving 95.1\%, 94.0\% and 90.1\% on three subsets. This indicates the beneficial effects of respective modules in our proposed network.

\begin{table}[h] \addtolength{\textwidth}{0.0cm} \renewcommand{\arraystretch}{1.1}
\begin{center}
\caption{Comparison of different settings on WIDER Face dataset.}
\label{tab3}
\setlength{\tabcolsep}{4mm}{
\begin{tabular}{|c|c|c|c|c|c|}
\hline
SBiFPN  & RFE & AM & Easy & Medium & Hard\\
\hline
$\checkmark$&  &  & 94.2\% & 93.6\% & 89.7\%\\
$\checkmark$&$\checkmark$ &  & 95.1\% & 93.9\%	& 89.8\%\\
$\checkmark$ & $\checkmark$ & $\checkmark$ & 95.1\%	& 94.0\% & 90.1\%\\
\hline
\end{tabular}}
\end{center}
\end{table}

\subsubsection{Performance of different backbones}\label{subsec434}
Table~\ref{tab4} shows the performance of our EfficientFace detector using different backbones. Since our network is inspired from EfficientDet \cite{r11}, we employ EfficientNet for backbone architecture with varying scaling factors involved, leading to different backbones for comparison. In our experiments, the same configuration is adopted for all the backbones. As expected, detection performance improves with the growing size of backbone. Compared to backbone EfficientNet-B0, the performance of EfficientNet-B5 is improved from 91.0\%, 89.1\%, 83.6\% to 95.1\%, 94.0\%, 90.1\% respectively on Easy, Medium and Hard subsets, providing dramatic performance boosts of 4.1\%, 4.9\% and 6.5\% with also approximately 10$\times$ growth in network parameters and MACs(G). This sufficiently indicates that the detector performance improves with the increase of model complexity to a large extent, whereas the model efficiency is severely compromised, which is consistent with the latest research results.

\begin{table}[h] \addtolength{\textwidth}{0.0cm} \renewcommand{\arraystretch}{1.1}
\centering
\begin{center}
\caption{Comparison of different backbones in both accuracy and efficiency on WIDER Face dataset.}
\label{tab4}
\setlength{\tabcolsep}{3.5mm}{
\begin{tabular}{|c|c|c|c|c|c|}
\hline
Backbone & Easy  & Medium & Hard & Params(M) & MACs(G)\\
\hline
EfficientNet-B0  & 91.0\%& 89.1\%& 83.6\%&3.94&4.80\\     
EfficientNet-B1  & 91.9\%& 90.2\%& 85.1\%&6.64&7.81\\     
EfficientNet-B2  & 92.5\%& 91.0\%& 86.3\%&7.98&10.49\\
EfficientNet-B3  & 93.1\%& 91.8\%& 87.1\%&11.53&18.28\\
EfficientNet-B4  & 94.4\%& 93.4\%& 89.1\%&19.36&32.54\\
EfficientNet-B5  & \textbf{95.1\%} & \textbf{94.0\%} & \textbf{90.1\%}&31.46&52.59\\
\hline
\end{tabular}}
\end{center}
\end{table}

\subsubsection{Comparison of EfficientFace with state-of-the-art detectors}\label{subsec435}
In this part, we compare EfficientFace with state-of-the-art detectors in terms of both accuracy and efficiency on WIDER Face validation set. As shown in Table~\ref{tab5}, the competing models involved in our comparative studies include not only heavy detectors such as MogFace, AlnnoFace and DSFD but also lightweight models like YOLOv5 variants and EXTD. In comparison to the heavy detectors, our EfficientFace model achieves competitive performance with significantly reduced parameters and computational costs. In particular, EfficientFace reports respective 95.1\%, 94.0\% and 90.1\% AP scores on Easy, Medium, and Hard subsets, which is on par with DSFD achieving 96.6\%, 95.6\% and 90.2\% accuracies. However, our model enjoys approximately $4\times$ reduced parameters and costs $6.5\times$ decreased MACs. Analogously, our EfficientFace is competitive with SRNFace-2100 with almost half of parameters and $4.8\times$ less MACs. Compared to the lightweight models, EfficientFace dramatically outperforms the competing methods. For example, our model reports 90.1\% on Hard subset, while exceeds YOLOv5n, EXTD-64 and LFFD by roughly 10\%, 5\% and 12\% in AP score. Although our network incurs more computational costs with more parameters compared to YOLOv5 and EXTD detectors, it inherits desirable efficiency from EfficientNet which serves as the building block of our proposed detector. Interestingly, as shown in Table~\ref{tab4}, when EfficientNet-B0 is used as the backbone in our EfficientFace detector, our model achieves competitive efficiency with lightweight YOLOv5n, while outperforming the latter by 3\% on Hard set. This sufficiently indicates that our model achieves a favorable trade-off between performance and efficiency.

\begin{table}[h] \addtolength{\textwidth}{-0.1cm} \renewcommand{\arraystretch}{1.1}
\centering
\caption{Comparison of EfficientFace and other advanced face detectors.}
\label{tab5}
\setlength{\tabcolsep}{2.7mm}{
\begin{tabular}{|c|c|c|c|c|c|}
\hline
Model & Easy  & Medium & Hard & Param(M) & MACs(G)\\
\hline
MogFace\_Ali-AMS\cite{r21} &94.6\%&93.6\%&87.3\%&36.07 & 59.11\\     
MogFace\_SSE\cite{r21} 	&95.6\%	&94.1\%	& - & 36.07	& 59.11\\     
MogFace\_HCAM\cite{r21} &95.1\%	&94.2\%	&87.4\%  & 41.79& -\\
MogFace-E\cite{r21}&\bfseries97.7\% &\bfseries96.9\% & 92.01\% & 85.67	& 349.14\\
MogFace\cite{r21}&97.0\%&96.3\%	&\bfseries93.0\%& 85.26	& 807.92\\
AlnnoFace\cite{r32}	&97.0\%	&96.1\%	& 91.8\%  & 88.01	& 312.45\\
SRNFace-1400\cite{r20}	&96.5\%	&95.2\%	& 89.6\%  & 53.38 & 251.94\\
SRNFace-2100\cite{r20}&96.5\%&95.3\%&90.2\%& 53.38& 251.94\\
DSFD\cite{r14}	&96.6\%	&95.7\%	& 90.4\%  & 120	  & 345.16\\
yolov5n-0.5\cite{r18}&90.76\%&88.12\%& 73.82\% & 0.45  & 0.73\\
yolov5n\cite{r18}	 &93.61\%&91.52\%& 80.53\% & 1.72  & 2.75\\
yolov5s\cite{r18}	 &94.33\%&92.61\%& 83.15\% & 7.06  & 7.62\\
yolov5m\cite{r18}	 &95.30\%&93.76\%& 85.28\% & 21.04 & 24.09\\
yolov5l\cite{r18}&95.78\%&94.30\%& 86.13\% & 46.60 & 55.31\\
EXTD-32\cite{r17}&89.6\%&88.5\%	& 82.5\%  & 0.063 & 5.29\\
EXTD-48\cite{r17}&91.3\%&90.4\%	& 84.7\%  & 0.10  & 7.7\\
EXTD-64\cite{r17}&92.1\%&91.1\%	& 85.6\%  & 0.16  & 13.26\\
LFFD\cite{r41} 	&91.0\%	&88.1\%	& 78.0\%  & 2.15  & -\\
Ours	        &95.1\%	&94.0\%	& 90.1\%  & 31.46 & 52.59\\
\hline
\end{tabular}}
\end{table}

\subsubsection{Comprehensive Evaluations on the four Benchmarks}\label{subsec436}
\begin{figure}
\centering
\subfigure[AFW]{
\includegraphics[width=1.4in,height=1.1in]{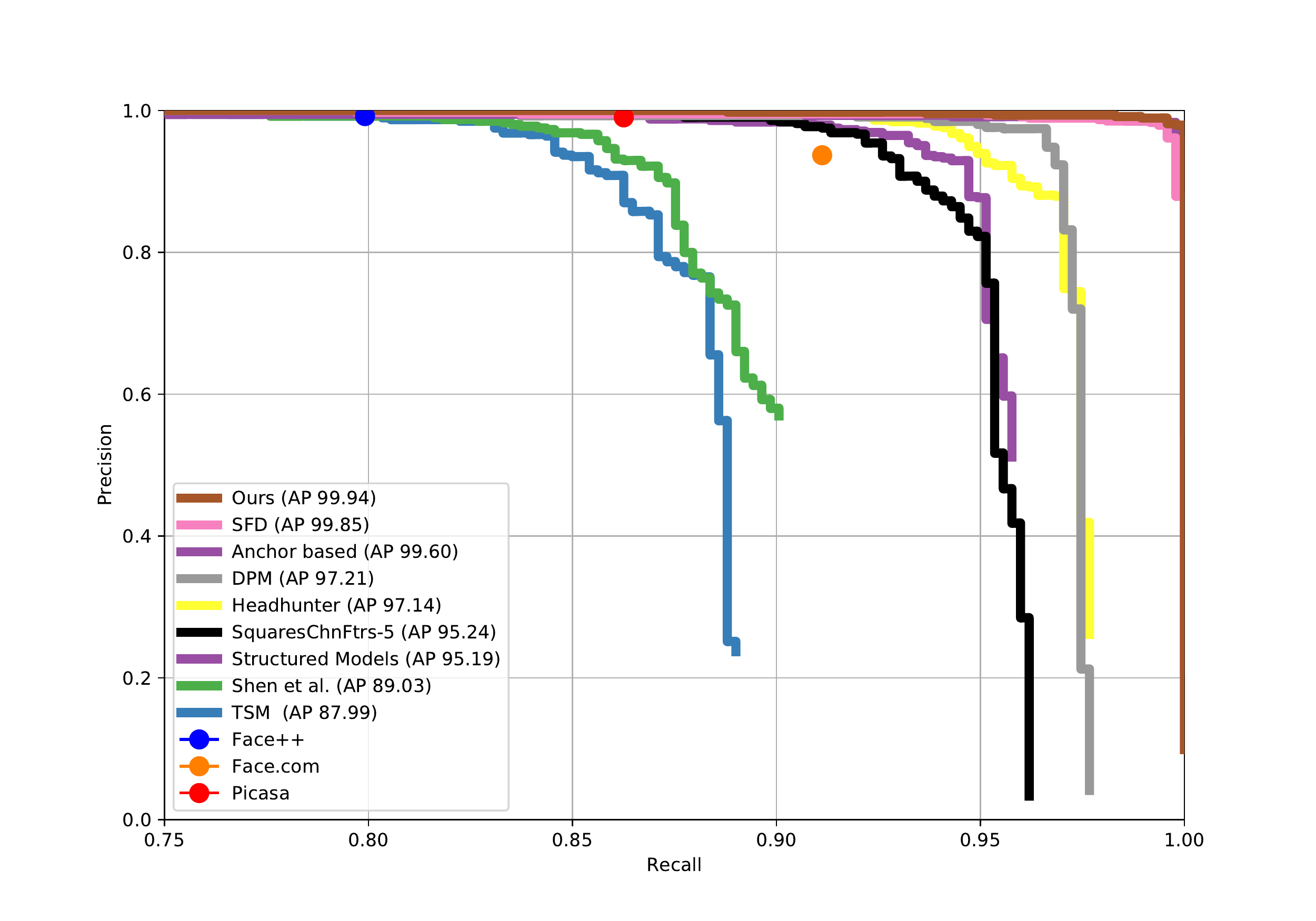}
}
\subfigure[PASCAL Face]{
\includegraphics[width=1.4in,height=1.1in]{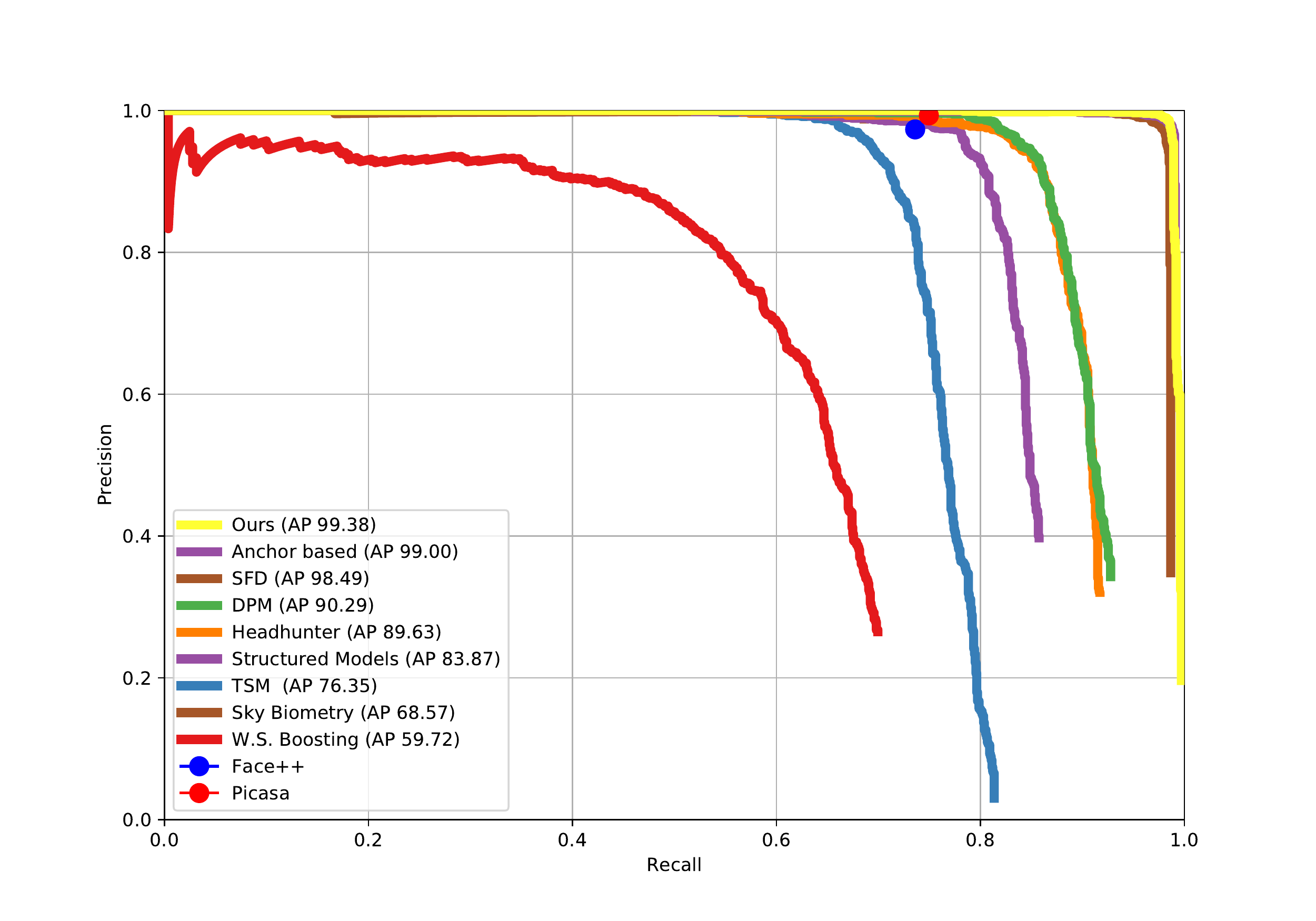}
}
\subfigure[FDDB]{
\includegraphics[width=1.4in,height=1.1in]{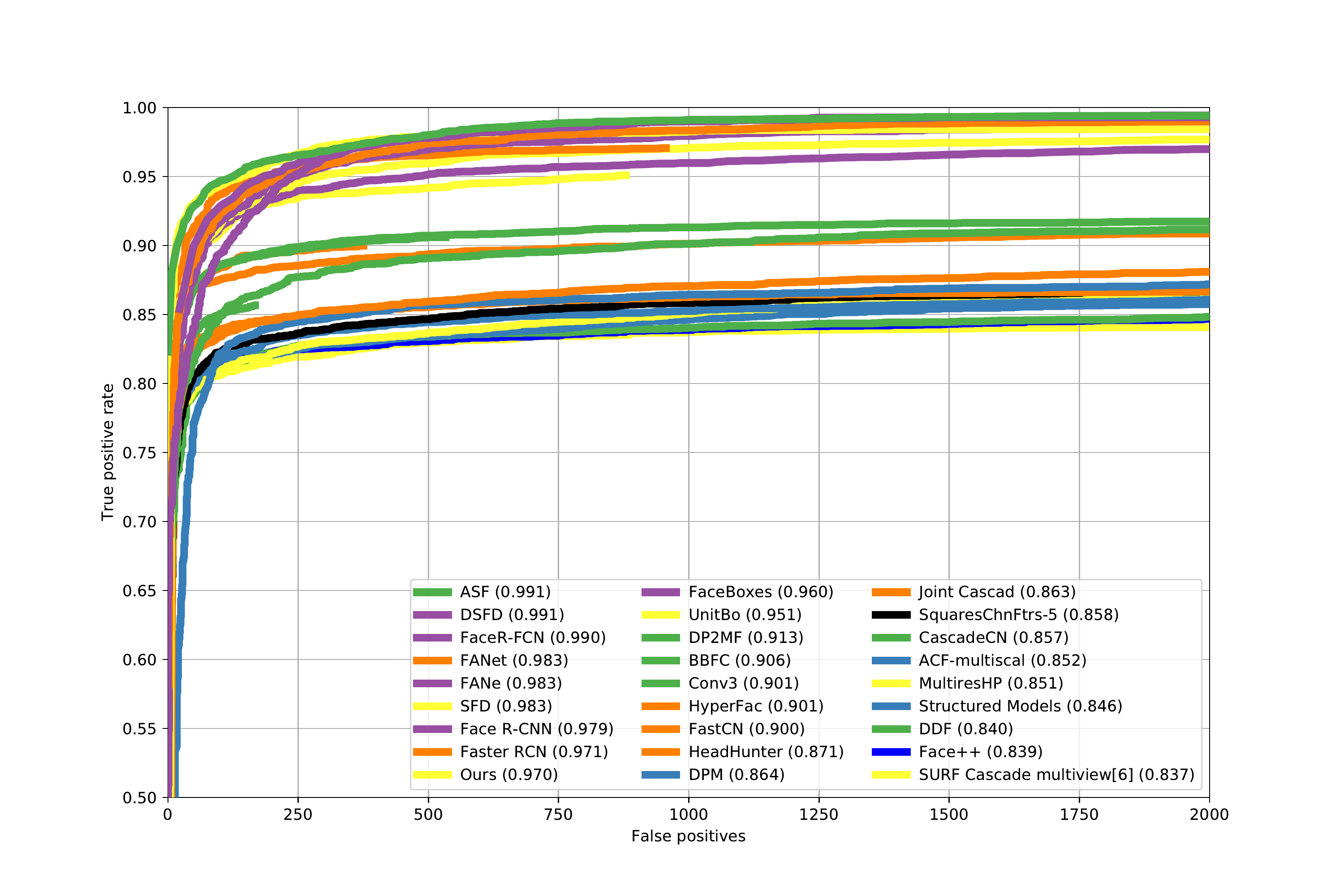}
}
\caption{Evaluation on common face detection datasets.}
\label{fig6}
\end{figure}

\begin{figure}
\centering
\subfigure[Val:easy]{
\includegraphics[width=1.4in,height=0.9in]{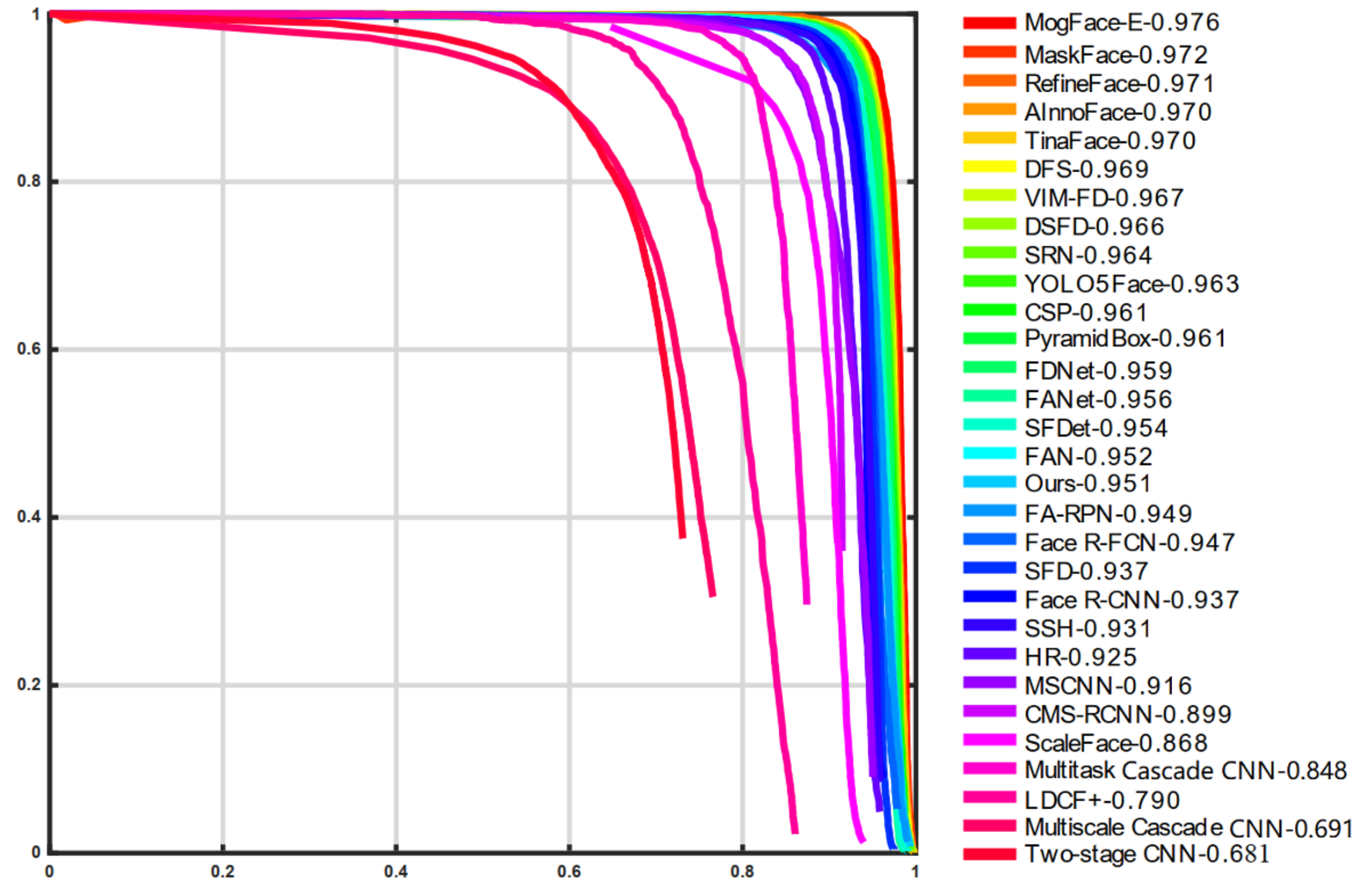}
}
\subfigure[Val:medium]{
\includegraphics[width=1.4in,height=0.9in]{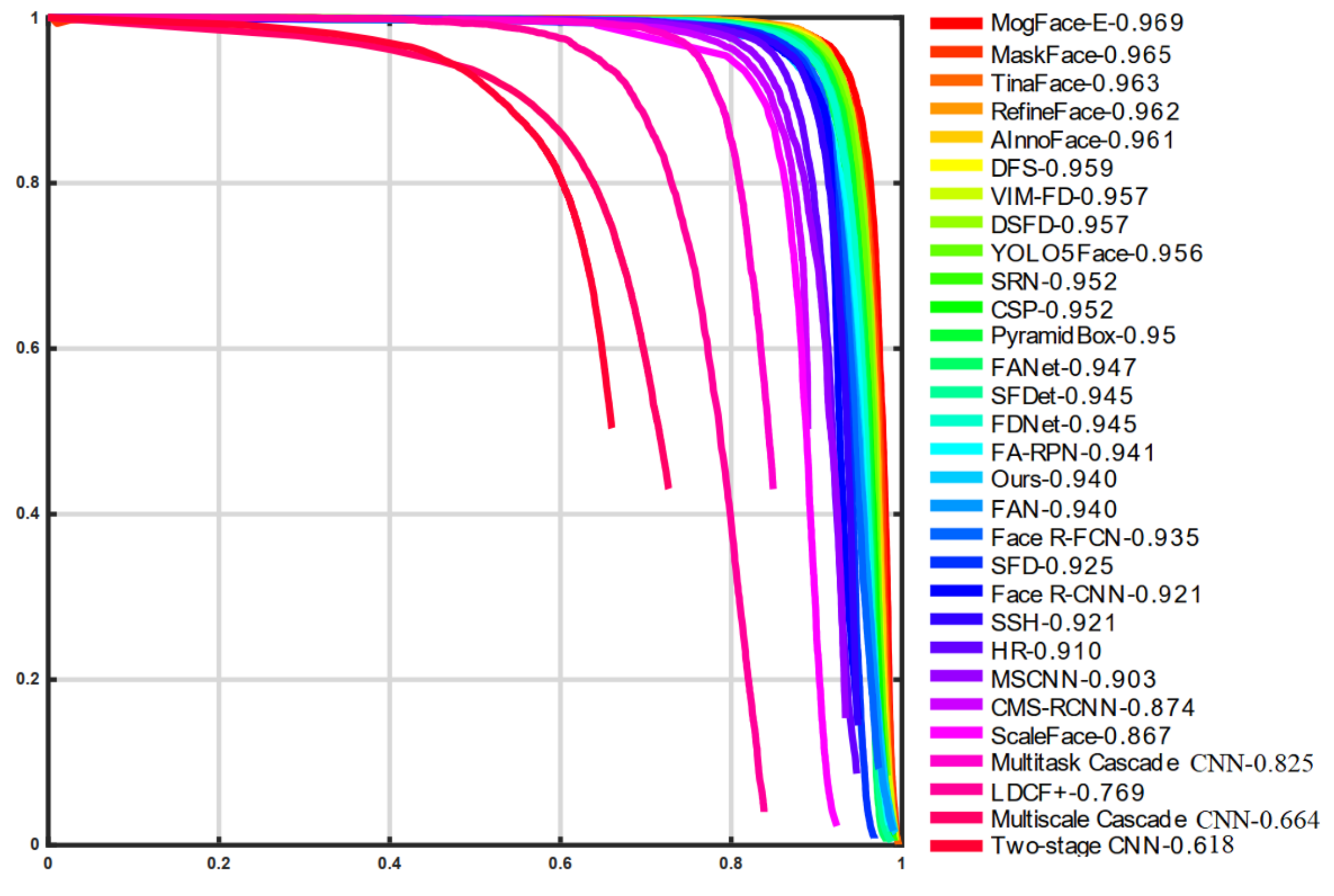}
}
\subfigure[Val:hard]{
\includegraphics[width=1.4in,height=0.9in]{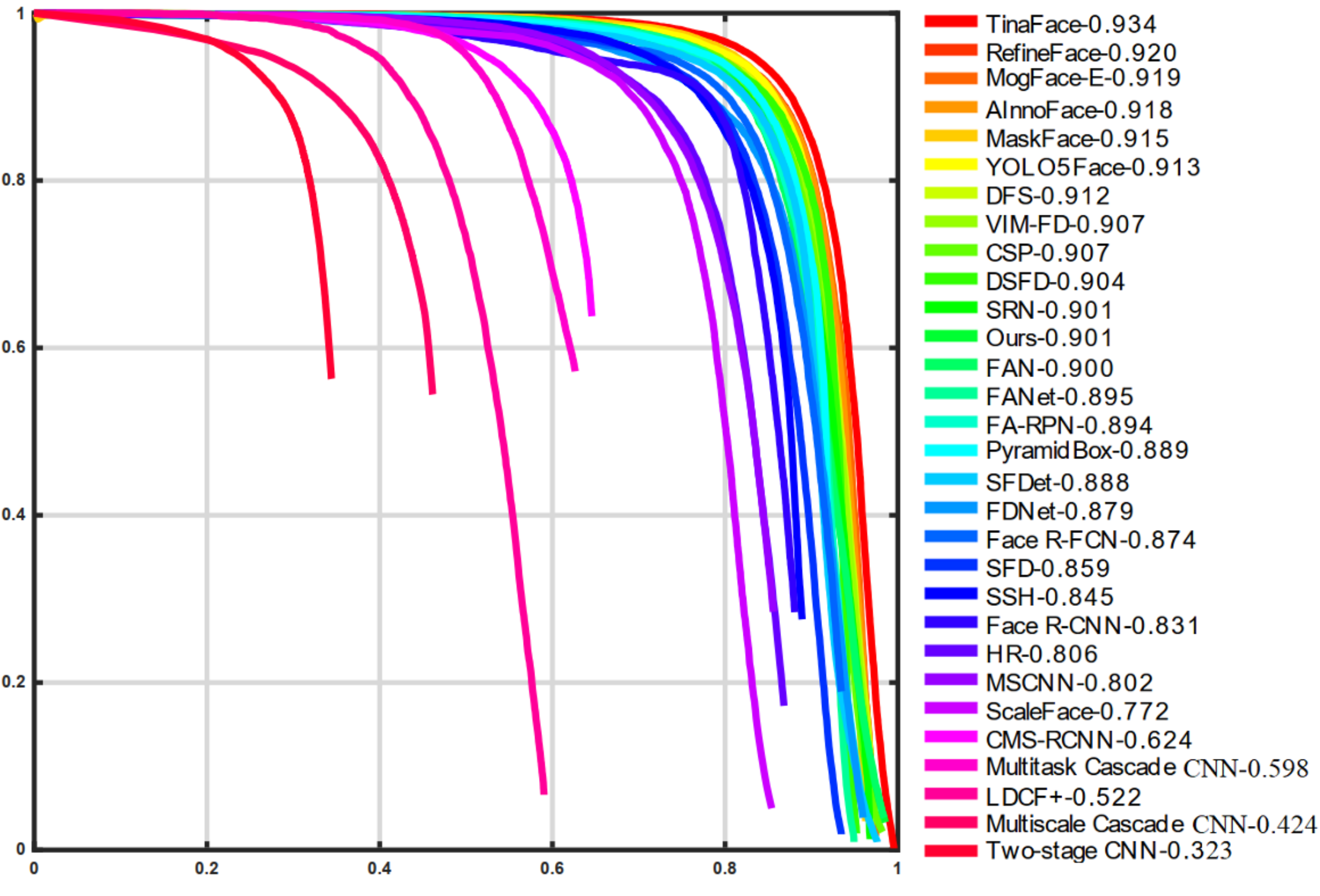}
}
\subfigure[Test:easy]{
\includegraphics[width=1.4in,height=0.9in]{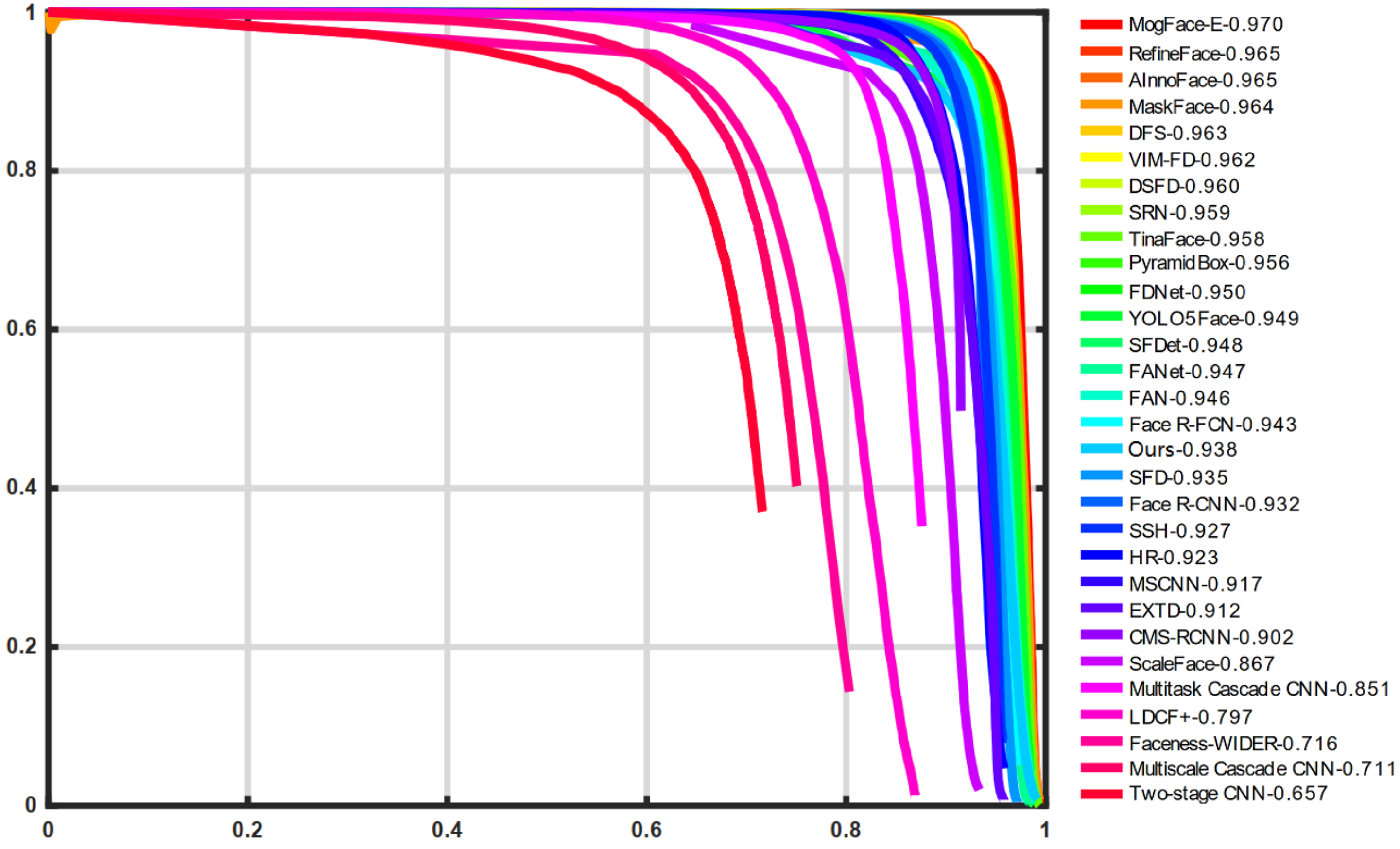}
}
\subfigure[Test:medium]{
\includegraphics[width=1.4in,height=0.9in]{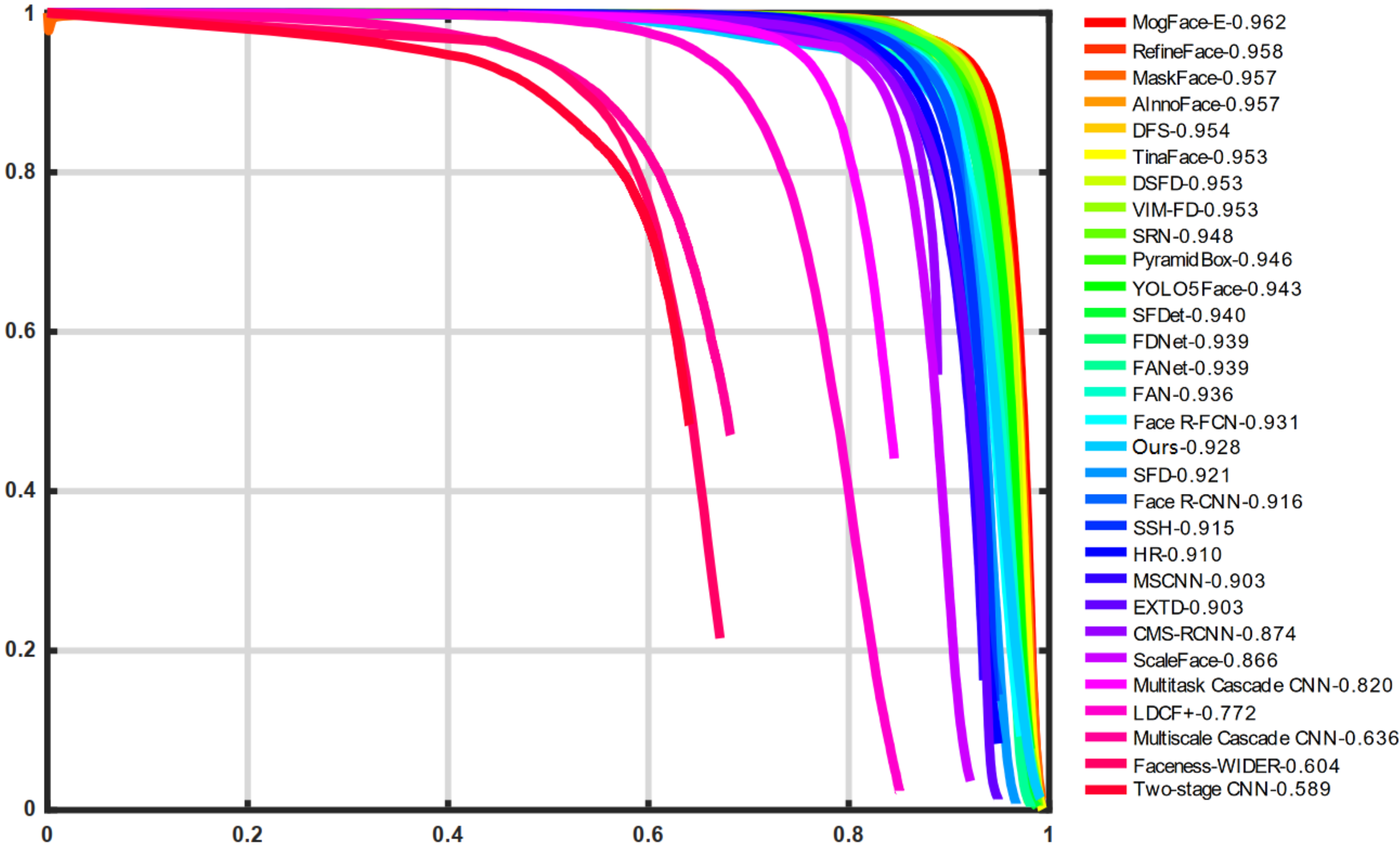}
}
\subfigure[Test:hard]{
\includegraphics[width=1.4in,height=0.9in]{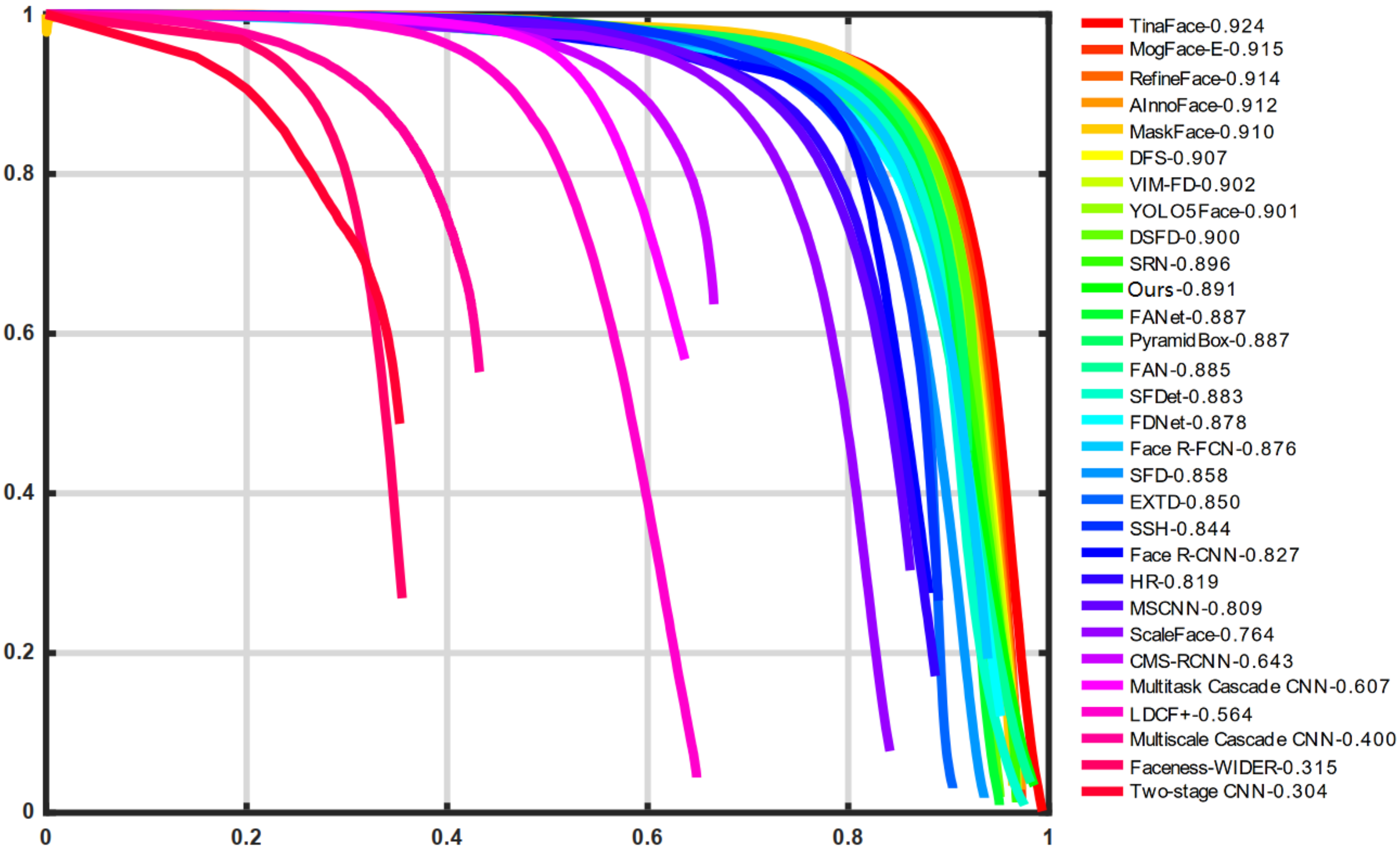}
}
\caption{PR curves of different methods on both validation and test set of WIDER Face dataset.}
\label{fig7}
\end{figure}

In this section, we will comprehensively evaluate EfficientFace and the other competing methods. In practice, our model is trained on training set of WIDER Face and then tested on the four benchmarks respectively. Fig.~\ref{fig7} demonstrates precision-recall (PR) curves achieved by different methods on both validation and test set of WIDER Face dataset. Although EfficientFace is still inferior to some advanced heavy detectors, it still achieves overall competitive performance with promising model efficiency. In addition to WIDER Face dataset, we also evaluate our EfficientFace on the other three datasets and carry out comparative studies. As illustrated in Table~\ref{tab6}, EfficientFace achieves respective 99.94\% and 99.38\% AP scores on AFW and PASCAL Face datasets. In particular, EfficientFace consistently beats the other competitors including even heavyweight models like MogFace and RefineFace~\cite{r49} on AFW. In addition, EfficientFace achieves competitive performance on PASCAL Face which is slightly inferior to RefineFace and FA-RPN. In addition to AP scores, we also provide PR curves of EfficientFace and other advanced detectors on AFW, PASCAL Face and FDDB datasets as shown in Fig.~\ref{fig6}. On AFW and PASCAL Face datasets, EfficientFace exhibits superior performance and consistently outperforms the other methods. On FDDB dataset, our model reports the true positive rate up to 97.0\% when the number of false positives is 1,000, which beats most of the face detectors and lags behind current state-of-the-art detector ASF by 2.1\%. Considering the model size and the computational costs of our network, EfficientFace reveals its promise in efficient face detection tasks.

\begin{table}[h] \addtolength{\textwidth}{0.0cm} \renewcommand{\arraystretch}{1.1}
\begin{center}
\caption{AP of EfficientFace and other detectors on the AFW and PASCAL Face dataset.}
\label{tab6}
\setlength{\tabcolsep}{6mm}{
\begin{tabular}{|c|c|c|}
\hline
	   Models & AFW	& PASCA Face\\
\hline
RefineFace~\cite{r49} & 99.90\%	& \bfseries99.45\%\\
FA-RPN\cite{r54} 	  & 99.53\%	& 99.42\%\\
MogFace\cite{r21}	  & 99.85\%	& 99.32\%\\
SFDet\cite{r53}       & 99.85\%	& 98.20\%\\
SRN\cite{r20}	      & 99.87\%	& 99.09\%\\
FaceBoxes~\cite{r52}  & 98.91\%	& 96.30\%\\
HyperFace-ResNet~\cite{r51}	& 99.40\%	& 96.20\%\\
STN~\cite{r50}	      & 98.35\%	& 94.10\%\\
Ours	              & \bfseries99.94\%	& 99.38\%\\
\hline
\end{tabular}}
\end{center}
\end{table}

\subsubsection{Qualitative Results}\label{subsec437}
\begin{figure}
\centering
\subfigure[Parade]{
\includegraphics[width=1.4in,height=2in]{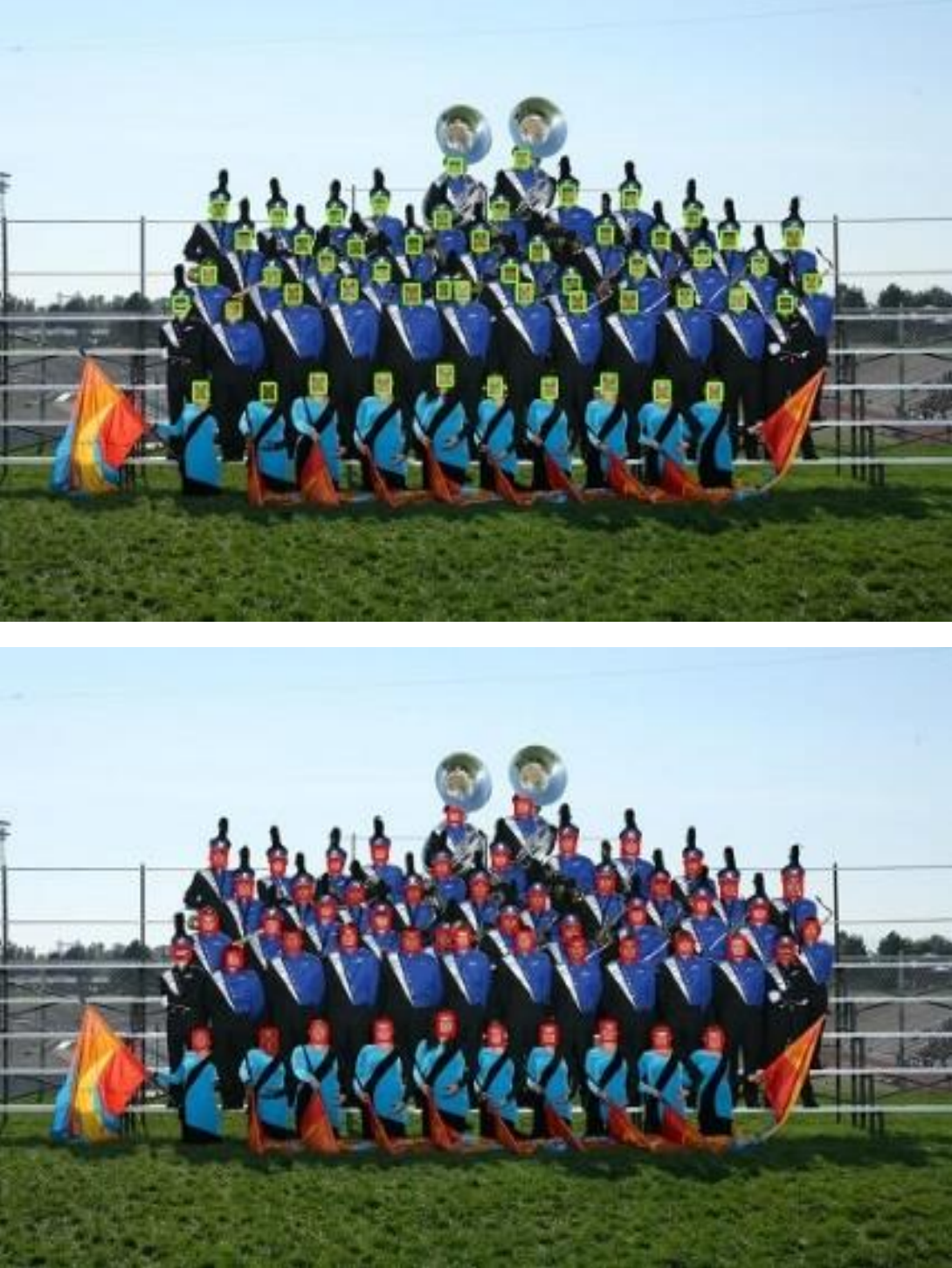}
}
\subfigure[Handshaking]{
\includegraphics[width=1.4in,height=2in]{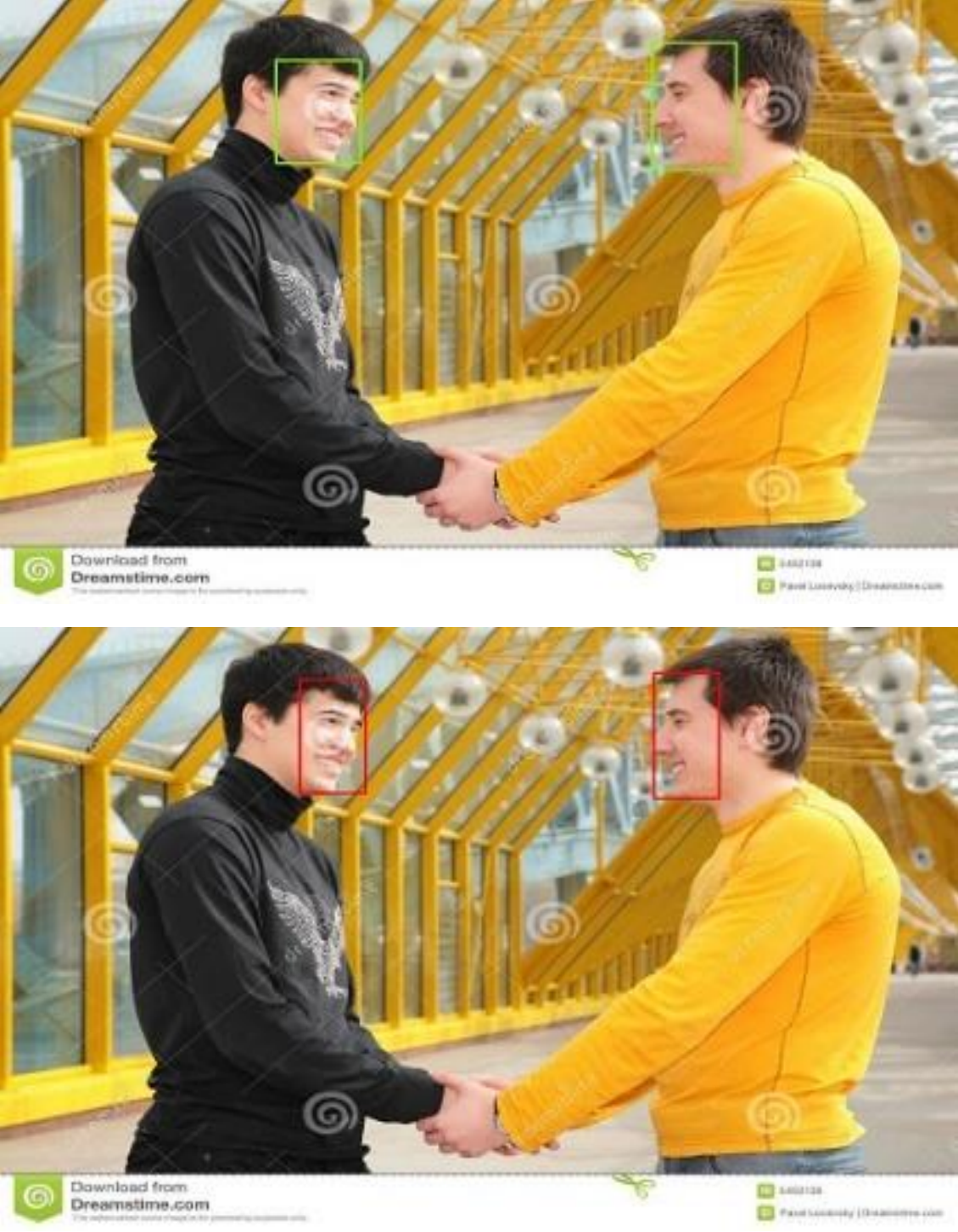}
}
\subfigure[Dancing]{
\includegraphics[width=1.4in,height=2in]{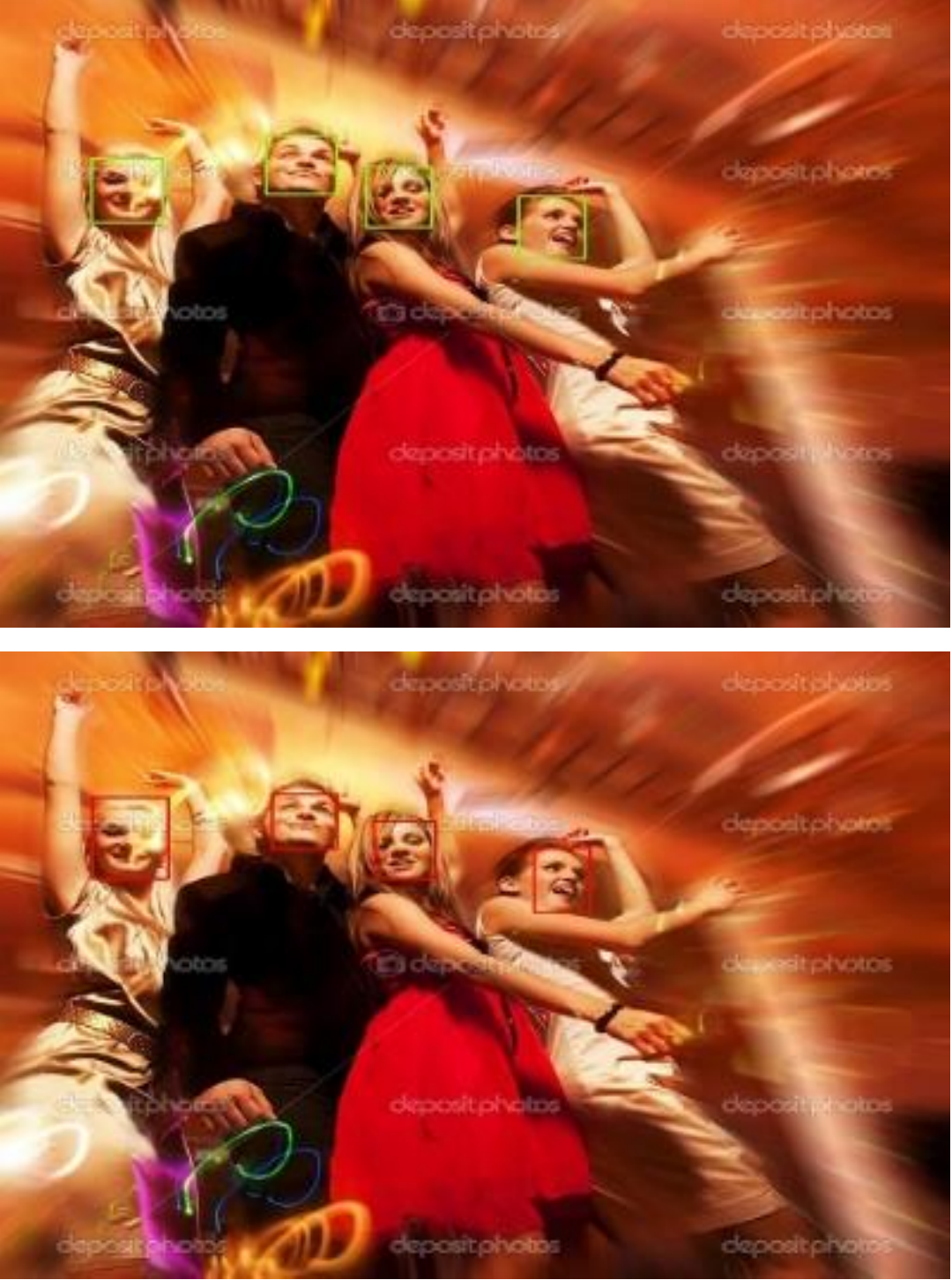}
}
\subfigure[Cheering]{
\includegraphics[width=1.4in,height=2in]{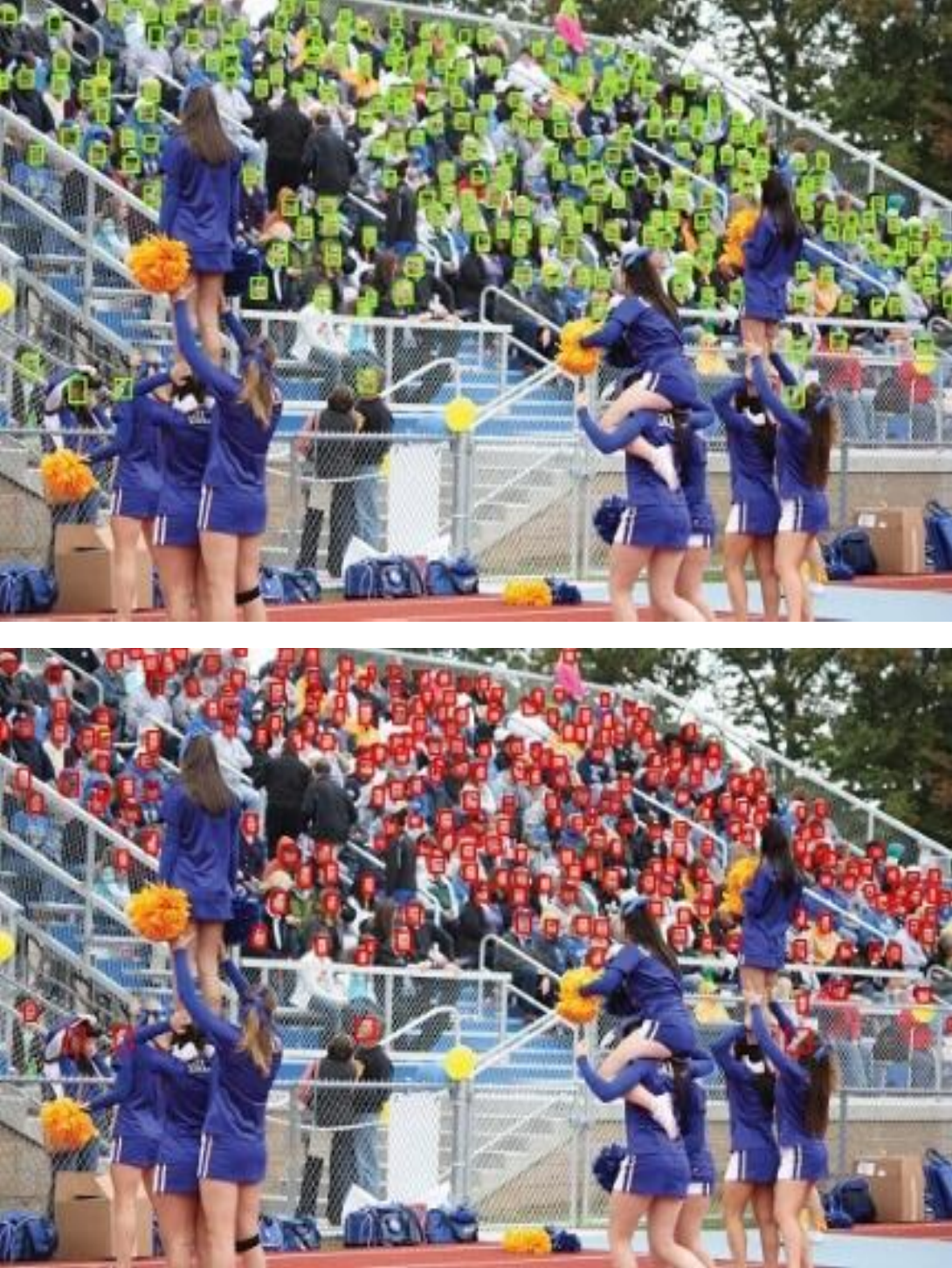}
}
\subfigure[Meeting]{
\includegraphics[width=1.4in,height=2in]{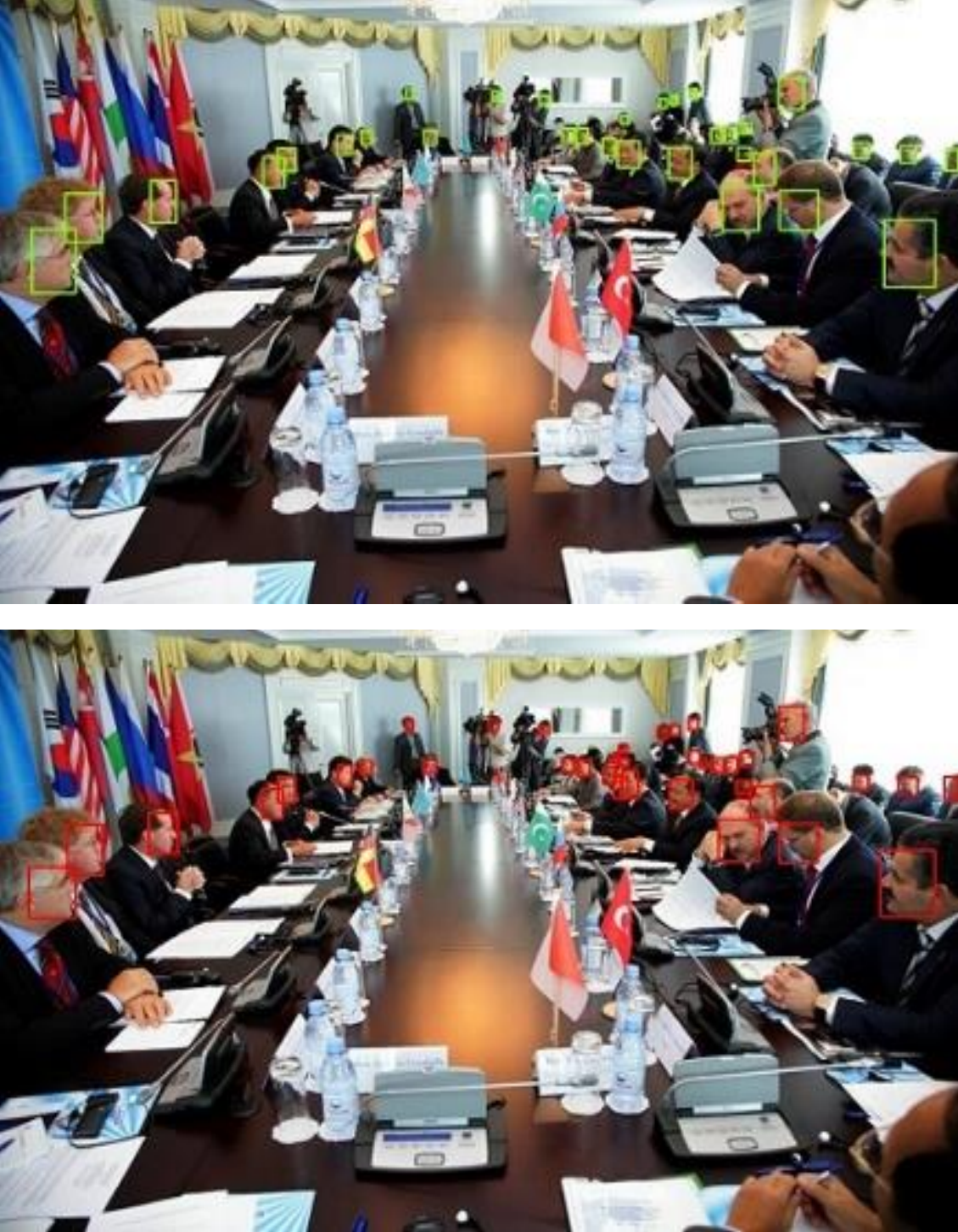}
}
\subfigure[Surgeons]{
\includegraphics[width=1.4in,height=2in]{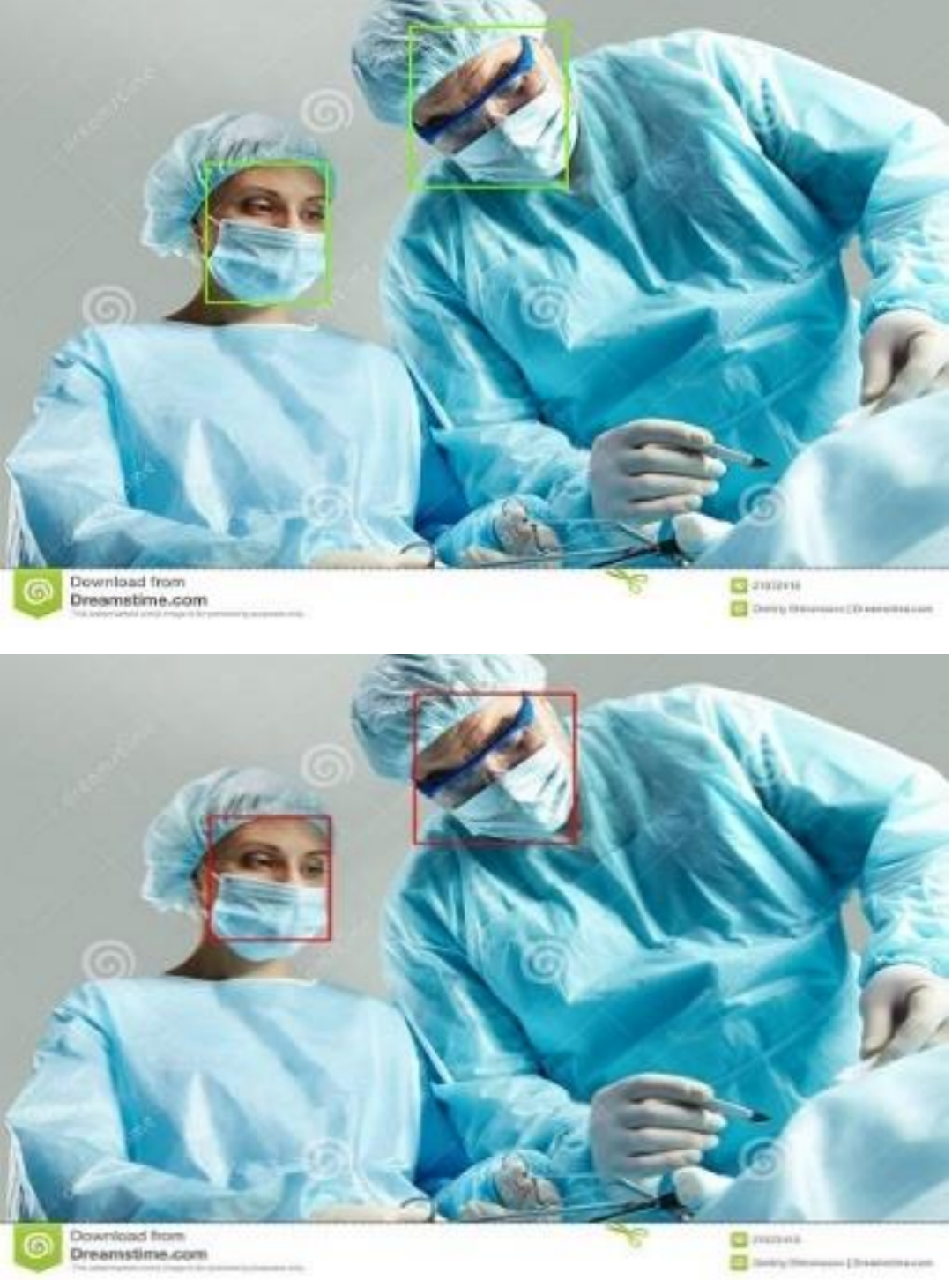}
}
\subfigure[Group]{
\includegraphics[width=1.4in,height=2in]{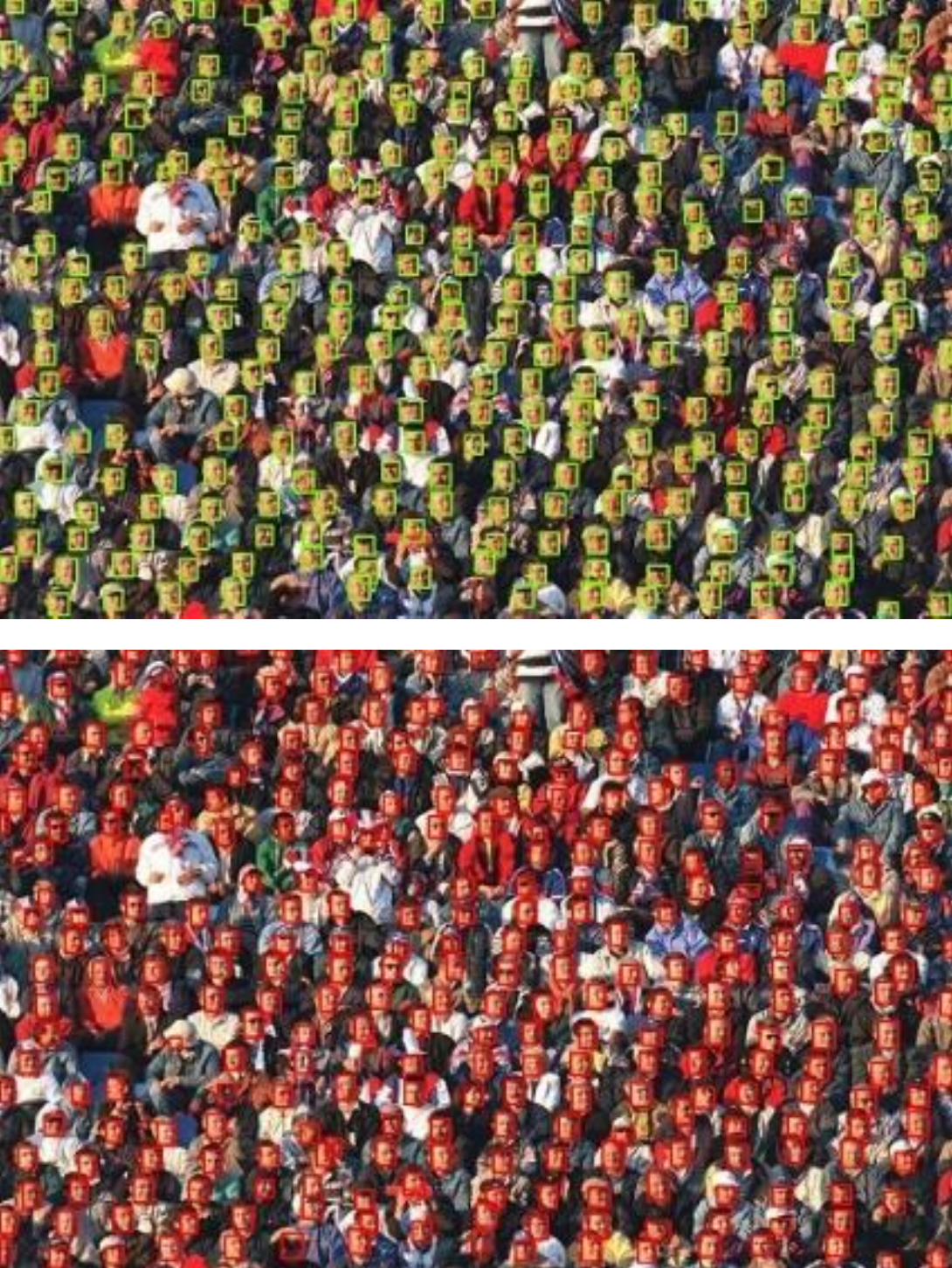}
}
\subfigure[Shoppers]{
\includegraphics[width=1.4in,height=2in]{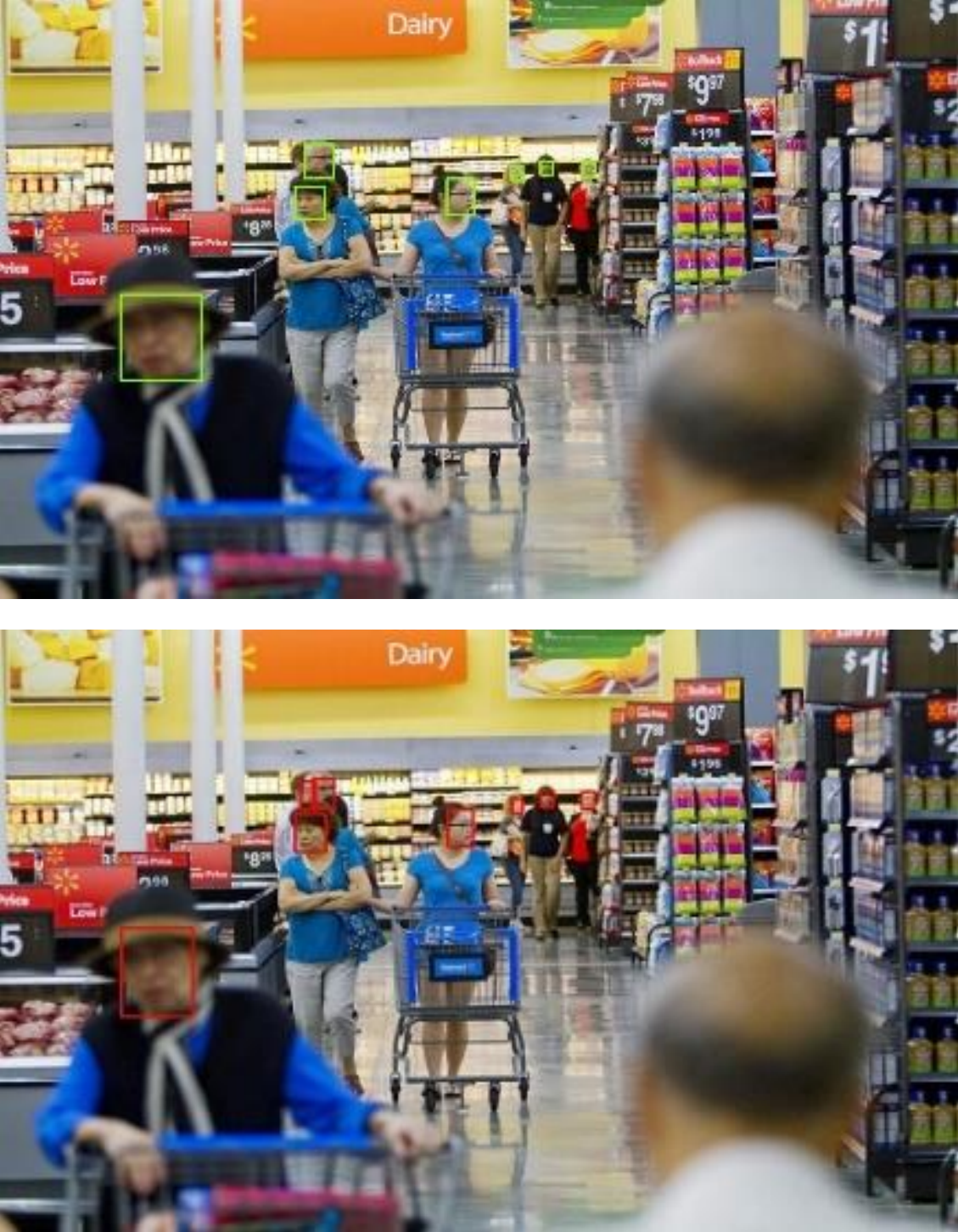}
}
\subfigure[Raid]{
\includegraphics[width=1.4in,height=2in]{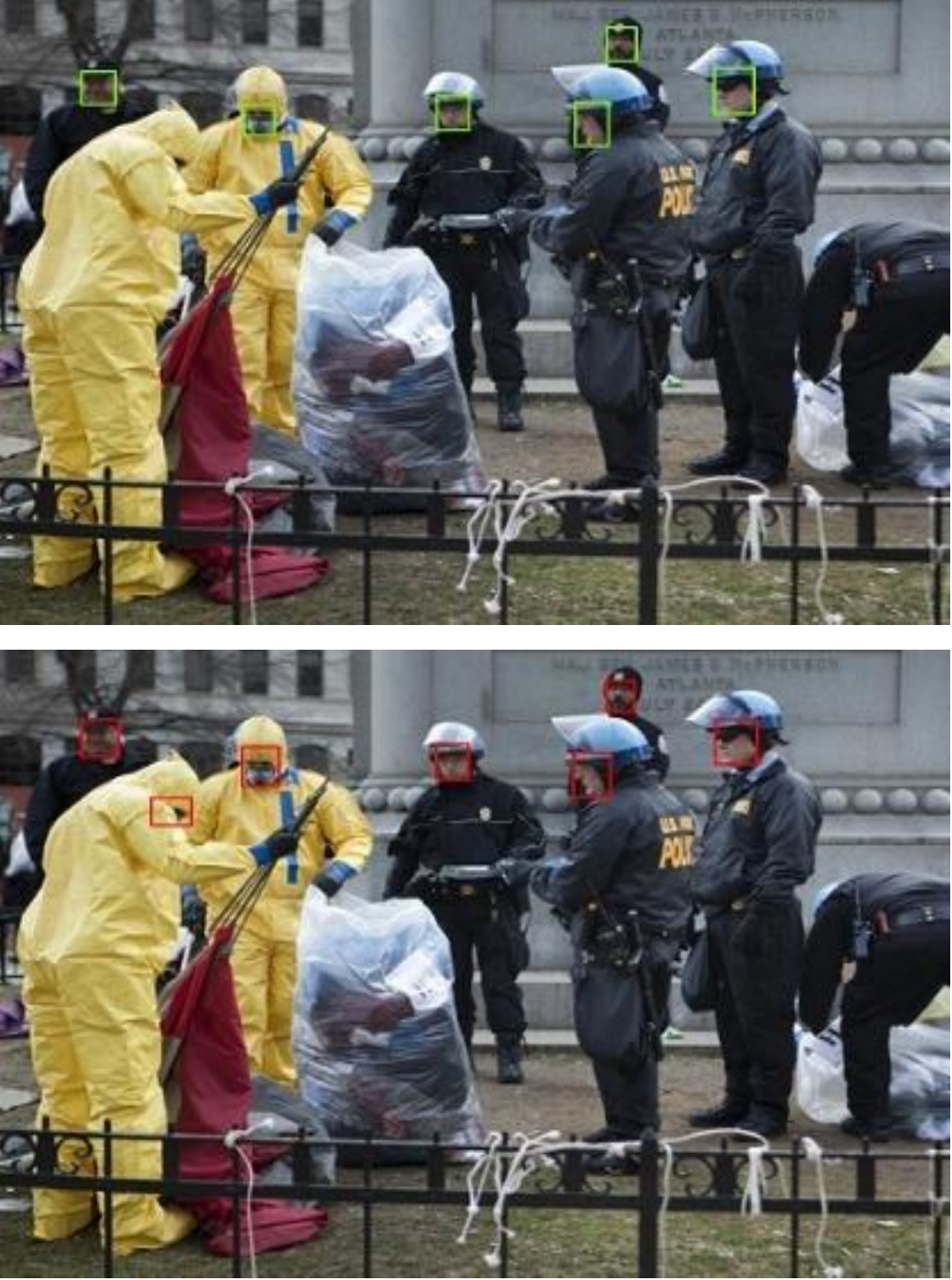}
}
\caption{Visualization of detection results achieved by EfficientFace in different scenarios. The detected faces are displayed in the first row with green boxes, while the corresponding ground truths are denoted with red boxes in the second row.}
\label{fig8}
\end{figure}

To intuitively demonstrate the performance of EfficientFace, we provide qualitative results of EfficientFace in various scenes as shown in Fig.~\ref{fig8}. The detected faces annotated in green boxes are displayed in the first row, while the corresponding ground truths are denoted with red boxes in the second row. The qualitative results suggest that our EfficientFace can accurately detect faces of various scales and well handle different challenges when massive cluttered faces are present, e.g., in cheering and meeting scenes. Thus, our method facilitates face detection in a variety of real-world scenarios.

\section{Conclusions}\label{sec5}
In this paper, we develop an efficient network architecture termed EfficientFace, which aims to improve the performance of lightweight face detectors due to their failure to deal with insufficient feature representation, faces with unbalanced aspect ratio and occlusion. Towards this end, we design a SBiFPN module to shorten the feature propagation pathway between low-level and high-level features and further strengthen feature expression by reusing characteristics. In addition, we add RFE module to detect faces with extreme aspect ratios in practical applications. Finally, attention modules including both spatial and channel attention are also incorporated in EfficientFace to better characterize the occluded faces. Our experiments on four public face detection datasets including AFW, PASCAL Face, FDDB and WIDER Face have demonstrated that our model achieves competitive performance compared to some advanced detectors, and reveals promising efficiency with reduced parameters and less computational costs. Thus, our method lends itself to the cases when both accuracy and efficient are demanding in practice.

\backmatter

\bibliography{ref}% common bib file



\end{document}